\documentclass[a4paper,fleqn]{cas-sc}
\usepackage{multirow} 
\usepackage{algorithm2e}
\usepackage{comment}
\usepackage[authoryear]{natbib} 
\usepackage{xcolor}
\usepackage{babel}
\DeclareUnicodeCharacter{2212}{-}
\usepackage{comment} 
\usepackage{hyperref}
\usepackage{graphics}
\usepackage{graphicx}
\usepackage{graphicx} 
\usepackage{microtype}
\usepackage{caption, subcaption} 
\def\tsc#1{\csdef{#1}{\textsc{\lowercase{#1}}\xspace}}
\tsc{WGM}
\tsc{QE}
\tsc{EP}
\tsc{PMS}
\tsc{BEC}
\tsc{DE}

\begin{document}
\let\WriteBookmarks\relax
\def\floatpagepagefraction{1}
\def\textpagefraction{.001}
\shorttitle{Offensive Language Identification}
\shortauthors{Hande et~al.}

\title [mode = title]{Offensive Language Identification in Low-resourced Code-mixed Dravidian languages using Pseudo-labeling}



\author[1]{Adeep Hande}[
                        orcid=0000-0002-2003-4836]
\ead{adeeph18c@iiitt.ac.in}
\credit{Conceptualization, Investigation, Methodology, Software, Visualization, Writing - original draft, Writing - review \& editing}

\address[1]{Indian Institute of Information Technology Tiruchirappalli, Tamil Nadu, India}
\cormark[1] 
\author[1]{Karthik Puranik}[style=indian,
                            orcid=0000-0002-5536-2258]
\ead{karthikp18c@iiitt.ac.in}
\cormark[1] 
\credit{Investigation, Methodology, Formal Analysis, Software, Writing - Original draft, Writing- editing}
\author[1]{Konthala Yasaswini}[style=indian,
                                orcid=0000-0001-5845-4759]
\ead{konthalay18c@iiitt.ac.in}
\credit{Investigation, Methodology,  Visualization, Writing - Original draft}
\author[2]{Ruba Priyadharshini}[orcid=0000-0003-2323-1701] 
\ead{rubapriyadharshini.a@gmail.com}
\credit{Investigation, Supervision, Data curation, Writing- reviewing \& editing}
\author[3]{Sajeetha Thavareesan}[orcid=0000-0002-6252-5393]
\ead{sajeethas@esn.ac.lk}
\credit{Investigation, Supervision, Writing - reviewing \& editing}
\author[4]
{Anbukkarasi Sampath}[orcid=0000-0003-0226-8150]
\ead{anbu.1318@gmail.com}
\credit{Formal Analysis, Investigation, Supervision, Writing - original draft, Writing - reviewing \& editing}
\author[4]
{Kogilavani Shanmugavadivel}[style=indian,
orcid=0000-0002-0715-143X]
\ead{kogilavani.sv@gmail.com}
\credit{Supervision, Writing - reviewing \& editing}
\author[5]
{Durairaj Thenmozhi}[orcid=0000-0003-0681-6628]
\ead{theni_d@ssn.edu.in}
\credit{Supervision, Writing - reviewing \& editing}
\author[6]
{Bharathi Raja Chakravarthi}[orcid=0000-0002-4575-7934]
\cormark[2] 
\ead{bharathi.raja@insight-centre.org}
\credit{Conceptualization, Formal Analysis, Investigation, Supervision, Data curation, Writing - original draft, Writing - reviewing \& editing}
\address[2]{ULTRA Arts and Science College, Madurai, Tamil Nadu, India}
\address[3]{Eastern University, Sri Lanka}
\address[4]{Kongu Engineering College, Erode, Tamil Nadu, India}
\address[5] {Sri Sivasubramaniya Nadar College of Engineering, Tamil Nadu, India}
\address[6]{Insight SFI Research Centre for Data Analytics,  National University of Ireland Galway, Galway, Ireland} 
\cortext[cor1]{Equal Contribution} 
\cortext[cor2]{Corresponding Author}

\begin{abstract}
Social media has effectively become the prime hub of communication and digital marketing. As these platforms enable the free manifestation of thoughts and facts in text, images and video, there is an extensive need to screen them to protect individuals and groups from offensive content targeted at them. Our work intends to classify code-mixed social media comments/posts in the Dravidian languages of Tamil, Kannada, and Malayalam. We intend to improve offensive language identification by generating pseudo-labels on the dataset. A custom dataset is constructed by transliterating all the code-mixed texts into the respective Dravidian language, either Kannada, Malayalam, or Tamil and then generating pseudo-labels for the transliterated dataset. The two datasets are combined using the generated pseudo-labels to create a custom dataset called CM-TRA. As Dravidian languages are under-resourced, our approach increases the amount of training data for the language models. We fine-tune several recent pretrained language models on the newly constructed dataset. We extract the pretrained language embeddings and pass them onto recurrent neural networks. We observe that fine-tuning ULMFiT on the custom dataset yields the best results on the code-mixed test sets of all three languages. Our approach yields the best results among the benchmarked models on Tamil-English, achieving a weighted F1-Score of 0.7934 while scoring competitive weighted F1-Scores of 0.9624 and 0.7306 on the code-mixed test sets of Malayalam-English and Kannada-English, respectively. The data and codes for the approaches discussed in our work have been released\footnote{\url{https://github.com/adeepH/Dravidian-OLI}}. 

\end{abstract}




\maketitle

\section{Introduction}

Social media has become a popular contrivance of communication in the 21st century and is the “democratisation of information” by converting people into publishers from the conventional readers \citep{unknownsocial}. 53.6\% of the world's population use social media \citep{bworld} which comprises a vivid structure between users from various backgrounds \citep{article23}. With its free expressing environment, it witnesses much content, including images, videos and comments from various age groups belonging to diverse regions, languages and interests. While the basic idea of social media remains to be communication and entertainment, users are seen using rude and defamatory language to express their views. Users might not appreciate such comments or posts and might be influential on teenagers. Offensive posts targeted on a group or an individual can lead to frustration, depression and distress \citep{article21, puranik2021iiitt}. Researchers recognised the need to detect and remove offensive content from social media platforms for a long period. However, there were a few challenges faced in this field. Though automating this process with the help of supervised machine learning models gave better accuracy than human moderators \citep{zampieri-etal-2020-semeval}, the latter were preferred as they could justify their decision in removing the comment/post from the platform \citep{risch-etal-2020-offensive}. Secondly, most of the comments and posts made were in code-mixed under-resourced languages \citep{chakravarthi-etal-2020-sentiment}. There was an absence of enough datasets and tools to produce state-of-the-art results to be implemented in these platforms. Our paper presents some unique approaches to give excellent F1 scores for code-mixed Dravidian languages, mainly Tamil, Malayalam, and Kannada.

Social media creates a whole new opportunity in the field of research. Non-English speakers tend to use phonetic typing, Roman scripts, transliteration, code-mixing and mixing several languages instead of Unicode\footnote{\url{https://amitavadas.com/Code-Mixing.html}}. Code-mixed sentences for Dravidian languages can be Inter-sentential which consists of pure Dravidian languages written in Latin script, Code-switching at a morphological level when it is written in both Latin and the Dravidian language and an Intra-sentential mix of English and the Dravidian language written in Latin script \citep{yasaswini-etal-2021-iiitt}. The foremost step in analysing code-mixed or code-switched data includes language tagging, which, if not accurate, can affect the results of other tasks. Language tagging has evolved over the years but is not yet satisfactory for analysing code-mixed data \citep{mandal-singh-2018-language}. The past years have seen enormous encouragement and research in code-mixing of under-resourced languages due to over-fitting. One of the main issues of dealing with code-mixed languages are the lack of annotated datasets and languages models being pretrained on code-mixed texts. The lack of code-mixed data resulted in constriction of data crisis, which affected the performance of various tasks. Transliteration of code-mixed data can increase the size of the input dataset. Transliteration refers to converting a word from one language to another while protecting the semantic meaning of the utterance and obeying the syntactic structure in the target language. The pronunciation of the source word is maintained as much as possible. While trying to get the critical features from a text or translating from one language to another, some language pairs like English/Spanish might not encounter any issue as \textit{Por favor} is written as \textit{Por favor} in English as they share the same Latin script. However, performing such tasks on Dravidian languages might pose problems, and transliteration can solve them to an extent \citep{DBLP:journals/corr/cmp-lg-9704003}. Supervised learning on small datasets or languages with limited resources can be complex. Thus, pseudo-labeling \citep{article-pseudo} can be employed to increase the performance considerably. In pseudo-labeling, the model is trained on labeled data to predict labels for a batch of unlabeled data. The predictions are then fed into the model as pseudo labelled data. 

\subsection{Research Questions}
In this paper, we attempt to address the following research questions:
\begin{enumerate}
    \item \textit{What architectures can be employed for effective cross-lingual knowledge transfer among code-mixed languages for offensive language identification?} \\
    We evaluate several recent approaches for offensive language identification, primarily focusing on cross-lingual transfer due to the persistence of code-mixed instances in the dataset. We have also used several state-of-the-art pretrained multilingual language models for three languages, Kannada, Malayalam, and Tamil.
    \item \textit{How do we break the curse of the lack of data for under-resourced Dravidian languages?}\\
    To overcome the barrier of lack of data, we revisit pseudo-labeling. We transliterate the dataset to the respective Dravidian language for our multilingual dataset and generate labels by using approaches such as pseudo-labeling. We combine the two datasets to form a larger dataset.
\end{enumerate}
\subsection{Contribution}
\begin{enumerate}
    \item We propose an approach to improve offensive language identification by emphasising more on constructing a bigger dataset, generating the pseudo-labels on the transliterated dataset, and combining the latter with the former to have extensive amounts of data for training.
    \item We experiment with multilingual languages models separately on the primary dataset, the transliterated dataset, and the newly constructed combination of both the datasets to examine if an increase in training data would improve the overall performance of the language models. We observe that this approach yields the best-weighted F1-Scores on all three languages concerning its counterparts.
    \item We have shown that our method works for three under-resourced languages, namely Kannada, Malayalam and Tamil, in a code-mixed setting. We also have compared our approaches to all other models that have been benchmarked on the datasets.
\end{enumerate}
The rest of our work is organised as follows. Section \ref{sec:1} talks about the related work on offensive language identification, while Section \ref{dravidian} entails a discussion about the Dravidian languages and their histories. Section \ref{Dataset} introduces the dataset used for the task at hand. Section \ref{method} discusses the several models and approaches to test their fidelity on the original code-mixed dataset, pseudo-labels procured for the transliterated data and the combination of the both. Section \ref{results} comprises a detailed analysis concerning the behaviour and results of the pretrained models when fine-tuned on code-mixed and transliterated data, and the results are compared with other approaches \citep{chakravarthi-etal-2021-findings-shared-task} from the shared task conducted by DravidianLangTech-2021\footnote{\url{https://dravidianlangtech.github.io/2021/}} at EACL 2021. We also perform the error analysis on the Kannada and Tamil predictions. Finally, Section \ref{conclusion} concludes our work and talks about potential directions for future work on Offensive Language Identification in Dravidian languages.
\section{Related Work}
\label{sec:1}
Recent years have witnessed a surge in the usage of offensive language on social media platforms. This surge has resulted in several corporations and organisations developing automated systems that filter offensive language on their platforms. 
In light of recent research on Dravidian languages, especially Dravidian code mixed text, a shared task for sentiment analysis of YouTube comments in Dravidian code mixed text has been presented \citep{10.1145/3441501.3441517}. The escalation of offensive language in social media has also resulted in some researchers conducting multi-disciplinary studies on the impact of offensive language on the mental health of its users \citep{chen2012detecting,xu2010filtering}. Early approaches to detect Offensive Language was very reliant on feature engineering applied to traditional machine learning classifiers \citep{dadvar2013improving}. More recent approaches to offensive language detection were hinging on Recurrent Neural Networks (RNNs) such as LSTMs, GRUs \citep{Pitsilis_2018} while employing word embeddings to RNNs to surpass the performance of traditional machine learning classifiers \citep{Bisht2020}. However, the success of transformer architecture led to its adaptation as the building blocks of encoder-decoder models, replacing the conventional Recurrent Neural Networks \citep{vaswani2017attention}. The success of pretraining has resulted in a pragmatic shift towards transfer learning for offensive language detection \citep{ruder2019transfer,liu2019nuli}. Offensive language detection is essentially treated as a sentence classification task \citep{risch-etal-2020-offensive}. Researchers have proposed an automated flame detection system that extracts features at a conceptual level to detect offensive language \citep{razavi2010offensive}. The terms hate speech, and offensive language is often misunderstood to mean the same thing, as hate speech is directed towards an entity. In contrast, offensive language is considered to be abusive and derogatory \citep{chaudhari2019}.

Several researchers have worked on developing systems to detect the offensive language in English, Turkish \citep{zampieri-etal-2020-semeval}. \citep{felbo-etal-2017-using} introduced the 'deepmoji' approach to elevate offensive language detection for English social media texts. This approach mainly depends on pretrain a neural network model for offensive language classification by utilizing emojis as weakly supervised training labels. In the work of \citep{6406271}, the authors suggested a Lexical Syntactic Feature architecture to strike a balance between identifying offensive content and possible offensive users in social media, arguing that, although current approaches consider messages as separate instances, the emphasis should be on the content's source. A topic-based mixture model integrated into the framework of semi-supervised training that takes advantage of a large amount of un-annotated Twitter data to detect offensive tweets was employed \citep{xiang-zhou-2014-improving}. The authors of \citep{10.1145/2396761.2398556} concentrated on Twitter and proposed a semi-supervised method along with statistical subject modelling for detection of offensive content.  

Toxic material is on the increase as digital knowledge becomes more widely available. As a result, detecting this form of language is extremely important. To overcome this issue, a combination of a state-of-the-art pretrained language model (CharacterBERT) and a conventional bag-of-words technique was used\citep{karimi2021uniparma}. Modern data science tools translate raw text into key features from which preprocessing or learning algorithms can make predictions for evaluating offensive communications, enabling for the identification and classification of toxic online comments \citep{noever2018machine}. Before the emergence of toxic text analysis, \citep{warner-hirschberg-2012-detecting} modelled hate speech as a word sense disambiguation issue in which SVM was used to classify data. To detect toxicity in social media messages, a technique based on a simple and powerful presumption: \emph{A post is at least as toxic as its most toxic span} \citep{xiang2021toxccin} was presented to boost the interpretability of transformers-based models like BERT and ELECTRA. The BiLSTM-CRF model was combined with Toxic Bert Classification to train the detection model \citep{luu2021uitisenlp} for recognizing toxic words from articles. The authors of \citep{brassard2019subversive} started by developing a sentiment detection tool that used a variety of lexicons and features such as word frequencies and negations, which proved that there is a strong correlation between sentiment and toxicity using this method. Later, they integrated sentiment data into a toxicity detection neural network and showed improved detection accuracy.

Hate speech is a form of offensive language that employs stereotypes to convey a hateful philosophy \citep{warner2012detecting}. \textit{Hate speech detection} is a closed-loop mechanism in which people know what is going on and deliberately avoid being detected. One of the challenges of automatic hate speech detection systems is the shifting of attitudes towards topics and historical context \citep{macavaney2019hate}. In the context of automated detection studies, offensive and abusive language are both used as overarching words for harmful content. Offensive language has a broader reach, and hope speech falls under each of these categories \citep{hande2021hope,yin2021generalisable}. The strong relationship between hate speech and actual hate crimes highlight the significance of identifying and moderating hate speech. Early detection of users who spread hate speech may lead to outreach efforts aimed at preventing the transition from speech to action \citep{waseem-hovy-2016-hateful}. The authors of \citep{malmasi2017detecting} used standard linguistic features and a linear SVM classifier to establish a baseline for discriminating between hate speech and profanity. \emph{Dialect} and \emph{race priming} was introduced, and conclusive proof that special consideration should be given to the conflicting effects of dialect in hate speech identification to prevent unintentional racial bias was found by the authors of \citep{sap-etal-2019-risk}. The authors of \citep{inproceedings} developed a pattern mining tool, PALADIN, to detect anti-social behaviours of users. PALADIN has demonstrated a method for information exploration, focusing on the disturbance in digital social networks using patterns. Work in \citep{burnap2014hate} describes a supervised machine learning text classifier that was trained and tested to differentiate between hateful and antagonistic responses based on race, ethnicity, and religion. By comparing classification results obtained from training on expert and amateur annotations, the authors of \citep{waseem2016you} investigated the impact of annotator knowledge of hate speech on classification models. Numerous challenges relate to detecting negative online behaviour, including identification that goes beyond simply recognizing offensive terms. \citeauthor{unsvaag2018effects} studied the potential and effects of including user features in hate speech classification, with an emphasis on Twitter, to close this distance.

To tackle Offensive language identification in Dravidian languages, several manually annotated datasets were constructed for Tamil \citep{chakravarthi-etal-2020-corpus}, Malayalam \citep{chakravarthi-etal-2020-sentiment}, and Kannada \citep{hande-etal-2020-kancmd} for sentiment analysis and offensive language identification. \citep{hande-etal-2020-kancmd} scraped the datasets from the comments section of the YouTube videos. \citep{hande2021benchmarking} benchmarked multi-task learning of supplemental tasks in under-resourced Dravidian languages. In multilingual countries like India, where the speakers are polyglots, it is evident that one would find the presence of code-mixed sentences, as the videos were scraped from social media. \citeauthor{chakravarthi-etal-2021-findings-shared-task} carried a shared task on Offensive language identification in Dravidian languages of Tamil, Malayalam, and Tamil for the user-generated comments. \citeauthor{chakravarthi-etal-2021-findings-shared-task} used DravidianCodeMix\footnote{\url{https://zenodo.org/record/4750858/}}, a multilingual code-mixed dataset manually annotated for sentiment analysis and offensive language identification \citep{chakravarthi2021dravidiancodemix}. The dataset consists of around 44,000 comments in code-mixed Tamil-English, 20,000 comments in Malayalam-English, and 7,700 comments in Kannada-English. The authors set the baselines using primitive machine learning algorithms for the datasets \citep{chakravarthi2021dravidiancodemix}, having a baseline weighted F1-Score of 0.65, 0.75, and 0.66 for the languages, in the said order. \citeauthor{jayanthi-gupta-2021-sj} devised an approach of task-adaptive pretraining multilingual BERT for offensive language identification that achieved the best-weighted F1-Score of 0.75 for Kannada. \citeauthor{saha-etal-2021-hate} pretrain XLM-RoBERTa from scratch, and developed the ensemble \citep{hande2021domain} of three models; Convolutional Neural Network (CNN), fine-tuned XLM-RoBERTa, and the custom pretrained XLM-R, scoring a weighted F1-Score of  0.78 and 0.97 in Tamil and Malayalam respectively, achieving the best results.

Majority of the works on offensive language identification focus more on the aspect of the model improvements. In this paper, we propose an approach that aims to address the lack of annotated data for low-resourced Dravidian languages, by transliterating the existing code-mixed dataset CM-TRA to their respective languages, and generating labels on them using pseudo-labeling. This approach results in effective cross-lingual transfer when fine-tuning multilingual language models. We fine-tune recent state-of-the-art pretrained language models that are very effective for cross-lingual transfer \citep{kalyan2021ammus}. We conduct sufficient experiments on all three different datasets (DravidianCodeMix, Transliterated-DravidianCodeMix, and Custom dataset CM-TRA), and our experimental results indicate that most of the fine-tuned language models fare relatively better than its results on the former datasets, with ULMFiT yielding the best results in all three languages.

\section{Dravidian Languages}
\label{dravidian}
The Dravidian languages comprise about 80 types, and are spoken in and around South Asia, mainly in southern and central India and countries like Singapore and Sri Lanka \citep{article2003,chakravarthi-etal-2021-findings-shared}. These languages flourished from the Dravidian civilization of the Indus Valley civilization around 4500 years ago\citep{chen-kong-2021-cs-dravidianlangtech}. The first signs of Dravidian languages are affirmed as Tamil-Brahmi scripts on the walls of caves in Madurai and Tirunelveli districts of Tamil Nadu, India, in the 2nd century BCE. While Tamil remains the oldest language in India, Telugu, Kannada, Malayalam and other Dravidian languages are prominently spoken by over 21\% of India's population\footnote{\url{https://en.Wikipedia.org/wiki/Dravidian\_languages}}. Tamil, the official language of the Indian state of Tamil Nadu, the Union Territory Puducherry and the nations of Sri Lanka and Singapore, is known to be one of the few most extended surviving languages in the world \citep{stein_1977}. The oldest literature among the Dravidian languages, the Sangam literature, was discovered in Tamil over 2000 years ago \citep{article96}. Linguists claim that Tamil is derived from Proto-Dravidian as there have been shreds of evidence of Tamil written in Brahmi script inscribed on rocks and caves around the 2nd century BC. Even today, 55\% of the inscriptions discovered are found in Tamil language\footnote{\url{https://web.archive.org/web/20060518064346/http://www.hindu.com/2005/11/22/stories/2005112215970400.htm}} with records also discovered in Sri Lanka, Egypt and Thailand.  Popular to the contrary beliefs, The Tamil writing system has its writing systems and is not Abugida, Abjad, nor Alphabet system. Tamil has vowels, pure consonants, uyirmey and Aytam. (strictly speaking, kuRRiyalukaram and KuRRiyalikaram). Unlike any system globally, Tamil is the only language that defines the length (duration) of a 'letter'.
Furthermore, it does not recognize any pure consonant. In Abugida, it is "obligatory: to end in Vowel. No Aytam either. Tamil system should be referred to as Tamil System where ezuttu is defined (unlike any other system, even aksara does not specify the length of any rules as to how many consonants can cluster to combine with a vowel. Kannada, another Dravidian language, is spoken mainly in Karnataka and the southwestern regions of India. Written in Kadamba script, Kannada was used by prominent dynasties like Chalukya, Rashtrakuta, Vijayanagara and Hoysala. With a history of over 2500 years, Kannada is said to have subtle influences from Sanskrit, Prakrit and Pali \citep{steever2018tamil}. Kannada is historically classified into Old Kannada (450–1200 CE), Middle Kannada (1200–1700 CE), and Modern Kannada (1700 CE–present) \citep{rice1982history, narasimhachar1988history} and is presently spoken by over 56.9 million people. Malayalam, the official language in the Indian state of Kerala and the Union Territories Lakshadweep and Puducherry (Mahé), is said to be derived from Tamil directly and separated at around 9th century CE or from the Proto-Dravidian from which Tamil has also originated. Vatteluttu script is currently used to write Malayalam is descended from Grantha script and is similar to the Tigalari script\citep{sekhar1951evolution, asher2013malayalam, george1972western}. Various literary works from the period of 9th to 11th centuries have been found, \citep{Bright1999REA} of which \textit{Ramacharitam} is said to be the earliest. Currently spoken by over 35 million people, it is one of the major Dravidian languages.

\section{Dataset}
\label{Dataset}
We use the offensive language data from DravidianCodeMix \citep{chakravarthi2021dravidiancodemix, chakravarthi-etal-2020-corpus,chakravarthi-etal-2020-sentiment, hande-etal-2020-kancmd}. The data comprises of various code-mixed comments on movie trailers on YouTube in Tamil, Malayalam and Kannada languages. The dataset is divided into training, development, and test sets with similar distribution for the three languages. Each set contains five different labels for the Malayalam dataset, while Tamil and Kannada datasets have six different types of labels, including the "Offensive Targeted Insult Other" label. We can observe an enormous class imbalance in the dataset, with “Not offensive” occupying a major share and "Offensive Targeted Insult Other" having a negligible amount of sample data for all three languages. The distribution of DravidianCodemix is tabulated in Table \ref{tbl1}. The class-wise distribution of the training and test set are tabulated in Table \ref{tbl2} and Table \ref{tbl3}. The six different classes include\citep{chakravarthi-etal-2021-findings-shared-task}:

\begin{itemize}
	\item \textbf{Not-Offensive (NO)}: Comments/post which is not impolite, rude and does not have obscenity, swearing, or profanity.
    \item \textbf{Offensive-Targeted-Insult-Individual OTI)}:Comments/ posts which are offensive and targeted at a particular person.
    \item \textbf{Offensive-Targeted-Insult-Group (OTG)}: Comments/ posts which are offensive and targeted at a group of individuals or community.
    \item \textbf{Offensive-Targeted-Insult-Other (OTO)}: Comments/ posts which are offensive but doesn't belong to any of the above two labels.
    \item \textbf{Offensive-Untargeted (OU)}: Comments/
    posts which are offensive but not targeting anyone.
    \item \textbf{Other Language (OL)}: Comments/
    posts are not in the intended language. 
\end{itemize}

\begin{table}[width=.6\linewidth,cols=4,pos=h]
\centering
\caption{Train-Development-Test Data Distribution}\label{tbl1}
\begin{tabular*}{\tblwidth}{@{} LRRRR@{} }
\toprule
Split & Kannada & Malayalam  & Tamil \\
\midrule
Training & 6,217 & 16,010 & 35,129 \\
Development & 777 & 1,999  & 4,388 \\
Test & 778 & 2,001 & 4392 \\ 
\midrule
Total & 7,772 & 20,010 & 43,909 \\
\bottomrule
\end{tabular*}
\end{table}

\begin{table}[width=.9\linewidth,cols=4,pos=h]
\caption{Class-wise distribution of the training set.}\label{tbl2}
\begin{tabular*}{\tblwidth}{@{} LRRRRRRR@{} }
\toprule
Languages/Classes & NO & OL & OTI  & OTG & OTO & OU & Total\\
\midrule
Tamil & 25,415 & 1,454 & 2,343 & 2,557 & 454 & 2,906 & 35,129 \\
Malayalam & 14,153 & 1,287 & 239 & 140 & - & 191 & 16,010 \\
Kannada & 3,544 & 1,522 & 487& 329 & 123 & 212 & 6,217 \\ 
\bottomrule
\end{tabular*}
\end{table}
\begin{table}[width=.9\linewidth,cols=4,pos=h]
\caption{Class-wise distribution of the test set}\label{tbl3}
\begin{tabular*}{\tblwidth}{@{} LRRRRRRR@{} }
\toprule
Languages/Classes & NO & OL & OTI  & OTG & OTO & OU & Total\\
\midrule
Tamil & 3,190 &165 & 315 & 288 & 71 & 368& 4,392 \\
Malayalam & 1,765 & 157 & 27 & 23 & - & 29 & 2,001 \\
Kannada & 417 & 185 & 75& 44 & 14 & 33 & 768 \\ 
\bottomrule
\end{tabular*}
\end{table}
\subsection{Code Mixing}
\begin{figure}[htbp]
    \centering
    \includegraphics[width=0.8\linewidth, height=6cm]{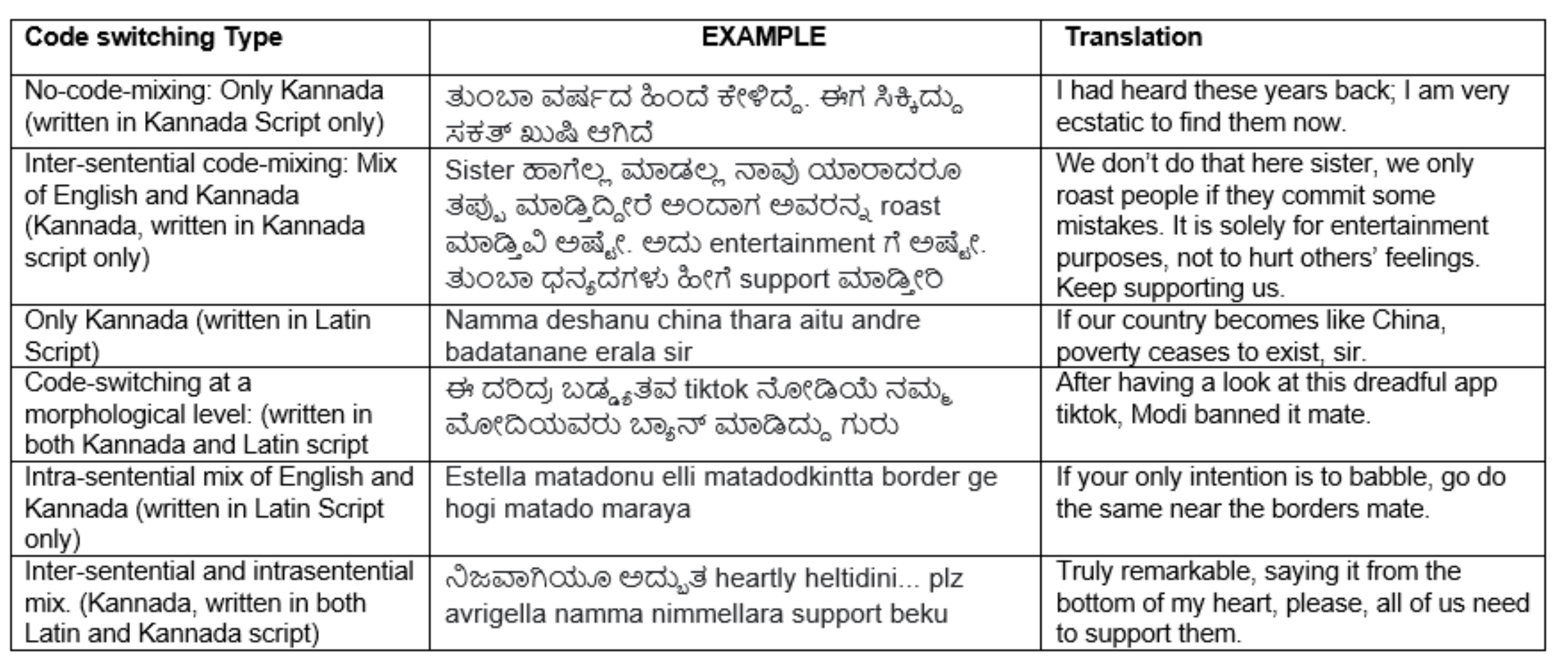}
    \caption{Example of code-mixing in the Kannada Dataset}
    \label{fig:kan_example}
\end{figure}

\begin{figure}[htbp]
    \centering
    \includegraphics[width=0.8\linewidth, height=8cm]{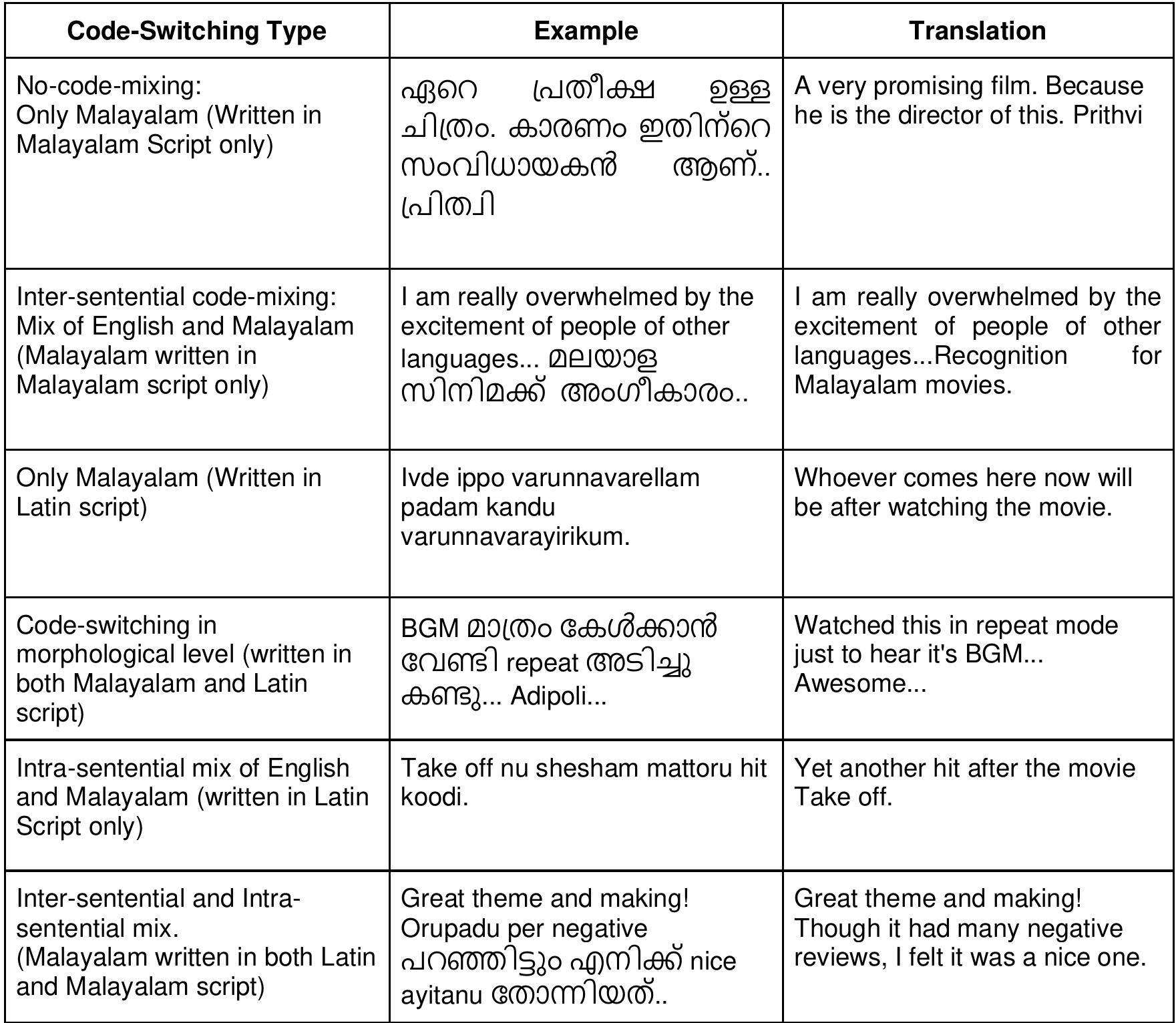}
    \caption{Example of code-mixing in the Malayalam Dataset}
    \label{fig:mal_example}
\end{figure}

\begin{figure}[htbp]
    \centering
    \includegraphics[width=0.8\linewidth, height=6cm]{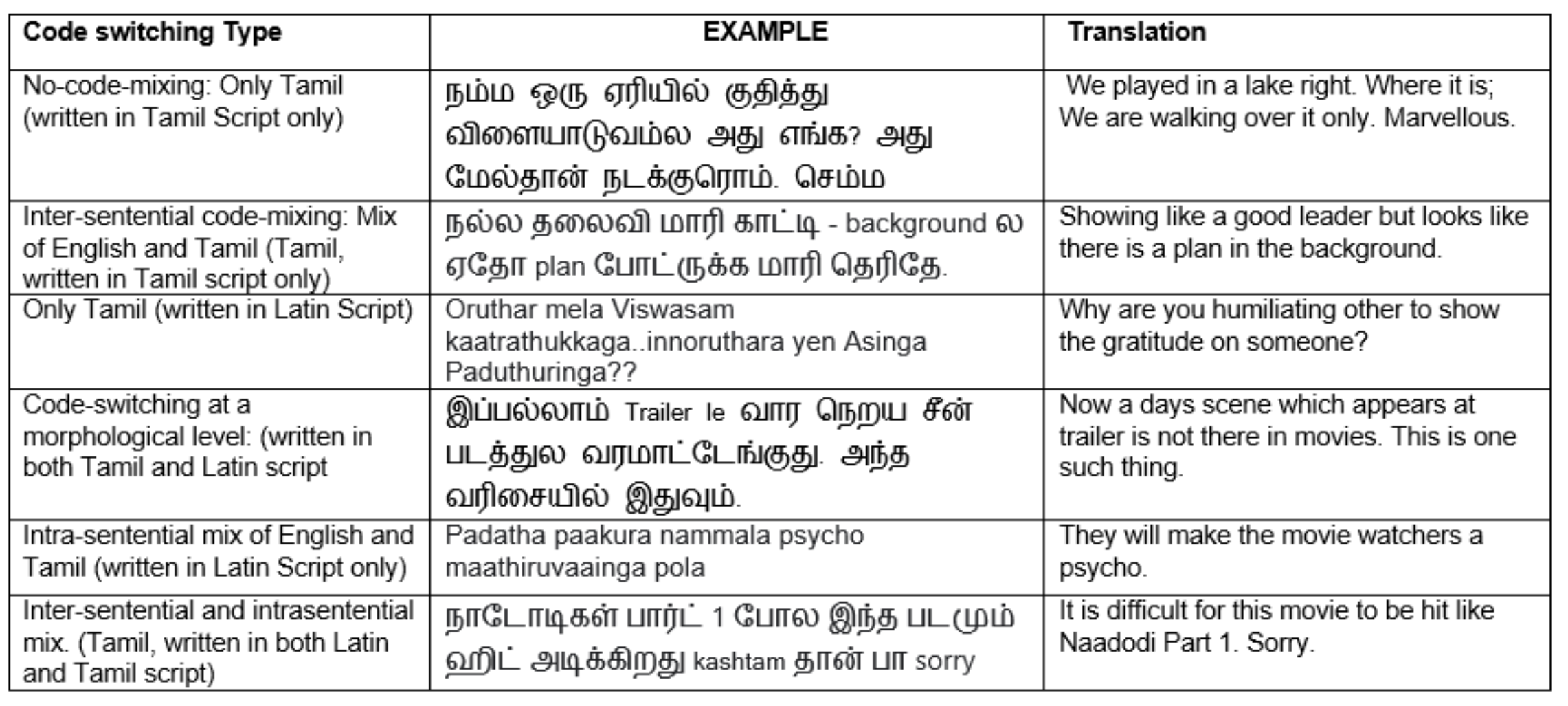}
    \caption{Example of code-mixing in the Tamil Dataset}
    \label{fig:tam_example}
\end{figure}

Multilingual speakers have a general trend of using distinct utterances from different languages, referred to as code-mixing. Code-mixing refers to the idea that a speaker switches from one language or variety to another in a text or a discussion. It is a prevalent phenomenon in a multilingual community \citep{chakravarthi-code-mix-survey}. Moreover, users like to blend various languages in their online platform interactions. There are different types of code-mixing in a language. We have given the examples of code-mixing in our corpora in Figure \ref{fig:kan_example}, Figure \ref{fig:mal_example}, and Figure \ref{fig:tam_example}.


\section{Methodology}
\label{method}
 
     
\subsection{Transformers}
Transformers is a revolutionary architecture in the field of Natural Language Processing introduced in the paper \cite{vaswani2017attention}. sequence-to-sequence tasks can be solved while managing long-range dependencies. Sequence-aligned RNN's or convolutions are not used as Transformers mainly depends on self-attention. The basic idea is to address the input and output with an attention mechanism and deliberately avoid recurrences. It executes this with the help of encoders and decoders \citep{9418446}. Attention takes in the input sequence and determines all the essential parts of the sequence.

The self-attention model lets the inputs interact with one another to find the representation of the sequence. For example, take the sentence "The car didn't cross the bridge as it was out of fuel". Humans might find it easy to interpret the sentence and map "it" with the car. However, that might not be the case with standard models. However, self-attention successfully maps them together. To calculate self-attention, we require query vector, key vector and value vector.
 
\begin{itemize}
    \item Query vector (Q): the current word 
    
    $Q=X*W_Q$.  
    \item Key vector (K): indexing method for value vector
    
    $K= X * W_K$.
    \item Value vector (V): data present in input word
    
    $V= X * W_V$.
\end{itemize}

Where, $W_Q$, $W_K$, and $W_V$ are the weight projections of Query, Key, and Value vectors respectively. Self-attention is parallelly and independently for each word in the Transformer's architecture. Once the outputs are concatenated and linearly transformed, we get the self-attention output for the required word. The process of calculating self-attention several times is referred to as multi-head attention.

\begin{align*}\label{eqn1}
    MultiHead(Q,K,V)=concat(head_1,head_2,....,head_n)W_O\\
    where, head_i=Attention(QW_i^Q,KW_i^K,VW_i^V)
\end{align*}

The equation represents the formula for multi-head attention.

\begin{equation}\label{eqn2}
    Attention(Q,K,V)=softmax(\frac{QK^T}{\sqrt{d_{k}}})V
\end{equation}
The equation \ref{eqn2} represents the formula for Self-attention for input matrices (Q, K, V).

\begin{figure}[htbp]
    \centering
    \includegraphics[width=\tblwidth, height=11cm]{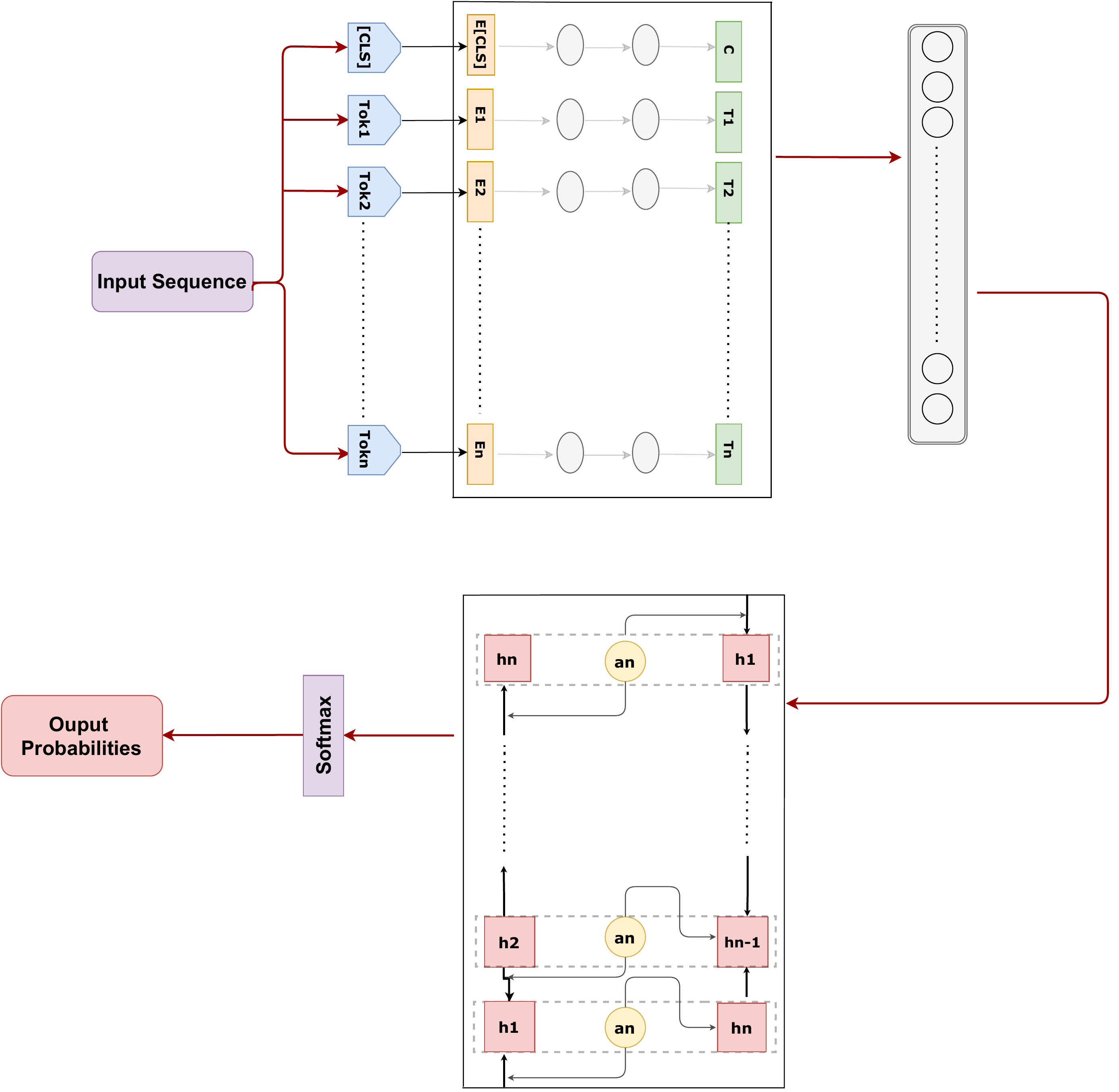}
    \caption{The architecture of feeding BERT-based Embeddings to a BiLSTM network.}
    \label{bert_bilstm}
\end{figure}

Furthermore, a thoroughly linked feed-forward network is present in the encoder and decoder layers, added independently to each of them. Two linear transformations with Rectified Linear Unit (ReLU) activation is comprised in between them. Though the linear transformations remain the same, the parameters differ from layer to layer. The equation of the Feed-Forward Network (FFN) is as follows:

\begin{equation}FFN(x) = \max(0,x * W_1, b_1)W_2 + b_2 \end{equation}
These linear transformation layers with the softmax function convert the output to predict the probabilities of the next token. The learned embeddings are employed to convert the input and output tokens into vectors with $d_{model}$ dimensions. Without RNNs and CNNs, the model should be made aware of the sequence of tokens and the position in a sequence. Positional encoding is added at the bottom of encoder stacks and decoder stacks of the input embeddings, and with dimension, $d_{model}$ which is the same as the embeddings, the two can be summed with ease. We compute the output vector's word embeddings and feed it into a bidirectional LSTM layer (BiLSTM) as shown in Figure \ref{bert_bilstm}. The pooler embedding \(T_E\) as input into the block. The model has three gates, namely, input gate $I_t$, output gate $O_t$ and forget gate $F_t$.The method of updating memory cell $C_i$ and current latency values $H_t$ is decided by three gates.For each node in LSTM, mathematical relationships between these gates are computed as follows:  
\begin{itemize}
\item $I_t = \sigma(w_i.[h_{t-1}, T_E] + b_i)$ 
\item $F_t = \sigma(w_f.[h_{t-1}, T_E] + b_f)$ 
\item $O_t= \sigma(w_o.[h_{t-1}, T_E] + b_o)$   
\item $C_i = \tanh(w_c.[h_{t-1}, T_E] + b_c)$  
\end{itemize}

\subsubsection{IndicBERT}
IndicBERT \citep{kakwani2020indicnlpsuite} is a multilingual ALBERT model that has been specifically trained on 12 major Indian languages. ALBERT was chosen as the base model because it has fewer parameters and is thus making it simpler to deploy and use in application areas. A single model was trained for multiple Indian languages to take advantage of their similarity. IndicBERT can support some of the under-represented languages. It is trained on IndicCorp and evaluated on IndicGLUE. 
One of the most extensive publicly accessible corpora for Indian languages is IndicCorp. It is created by exploring and filtering thousands of online outlets, mainly new books and magazines. To tokenize the sentences in each language, a sentencepiece tokenizer is trained with IndicCorp. This tokenized corpus trains a multilingual ALBERT using the traditional masked language model (MLM). To give low-resource languages a more significant representation, exponentially smoothed data weighting is implemented across languages. A vocabulary of 200k to fit various scripts and massive vocabularies of Indic languages is selected. The Indic General Language Understanding Evaluation Benchmark, IndicGLUE, is a robust Natural Language Understanding (NLU) benchmark introduced to extensively evaluate language models on multiple Indian languages. It consists of 6 tasks. After pretraining, IndicBERT is fine-tuned on the specific task in IndicGLUE using the corresponding training sets. For Sentiment Analysis, the last layer's representation of the [CLS] token is fed to a linear classifier with a softmax layer to determine a probability distribution over the categories. It is important to note that IndicBERT is much smaller than two of the most potent performing multilingual models: mBERT and XLM-R base, despite being trained on larger Indic language corpora.

\subsubsection{DistilBERT}
Operating large models in peripheral computational training or inference resources remains difficult, as Transfer Learning from large-scale pre-trained models becomes more extensive in Natural Language Processing (NLP). Although models like BERT and XLM-RoBERTa have millions of parameters, increase performance substantially, training much greater models results in compelling performance on downstream areas. This shift presents a multitude of challenges. DistilBERT \citep{sanh2020distilbert} is a light transformer model trained by distilling BERT base. It is a method for pretraining a smaller language representation model. It has 40\% lesser parameters than \emph{bert-base-uncased} and runs 60\%faster at the same time maintaining 97\% of its language understanding capabilities. Using the DistilBERT model pretrained with knowledge distillation resulted in a similar performance on numerous downstream tasks. 

\textit{Knowledge distillation} is a compression method in which a smaller model - the student - is trained to emulate the behaviour of a larger model - the teacher or an ensemble of models. The student has the same architecture as BERT, but the token-type embeddings and the pooler are removed from the architecture, and the number of layers is reduced by a factor of 2. The student is initialised by taking one layer out of two from the teacher through the advantage of the shared dimensionality between their networks. The student is trained with the triple loss, which is a linear combination of the distilled loss \emph{L\textsubscript{CE}}, the supervised training loss, which is the masked language modelling loss  \emph{L\textsubscript{MLM}} and a cosine embedding loss ( \emph{L\textsubscript{cos}}) as it matches the student and teacher hidden states vectors. The triple loss is mathematically defined as :

\begin{equation}
L_{ce} = \sum_{i} t_i * log(s_i)
\end{equation}

where \emph{t\textsubscript{i}} (resp. $s_i$ ) is a probability estimated by the teacher (resp. the student). This training loss leads to an upscale training signal by leveraging the complete teacher distribution.
\begin{table*}[htbp]
\begin{tabular}{lll}
\toprule
Hyper-parameters & Characteristics\\
\midrule
Number of LSTM Units &  256\\
Loss & Cross Entropy\\ 
Epoch & 5   \\
Batch size & [16, 32, 64]   \\
Optimiser &  AdamW \citep{loshchilov2019decoupled} \\
Dropout  &  0.4 \\
Learning rate & 2e-5\\
Max length & 128\\
\noalign{\smallskip}\hline
\end{tabular}
\caption{Various hyper-parameters used for our experiments}\label{params}
\end{table*}

\subsubsection{ULMFiT}
A language model (LM) determines the probability distribution over a sequence of words. When fine-tuned with a classifier, LMs encountered catastrophic forgetting and overfitted to small datasets. Due to the length of the vocabulary, statistical LMs suffer the effects of data sparsity. A novel approach, Universal Language Model Fine-tuning (ULMFiT) \citep{howard2018universal} is introduced that solves these challenges and promotes efficient inductive transfer learning for any NLP task. ULMFit pretrains a language model(LM) on a broad general-domain corpus and fine-tunes it on the target task using standard methods. It employs a single architecture and training process. No custom feature engineering or preprocessing is required in this model. It involves three stages: 
\begin{enumerate}
    \item \textbf{General-domain LM pretraining:}
    
    The language model(LM) is pretrained on WikiText \citep{merity2016pointer} comprising of 28,595 preprocessed Wikipedia articles 103 million words to attain the same level of accuracy as computer vision models trained on ImageNet corpus. Although this stage is the most expensive, it can capture the general language properties. After that, the pre-trained model may be used for further NLP applications.
    
    \item \textbf{Target task LM fine-tuning:}
    
    The target task dataset will have different distribution regardless of how varied the general-domain corpus is for pretraining the model. In this stage, the LM is fine-tuned on the target task dataset to learn its distinctions using discriminative fine-tuning and slanted triangular learning rates. Rather than utilising the same learning rate in the entire model, the authors propose to use different learning rates for each layer. It initially selects the learning rate of the last layer by fine-tuning only that layer and then applies the following formula to the lower layers -
    \[ \eta^{I−1} = \frac{\eta^I}{2.6},  \textrm{where} \ \eta^I \ \textrm{is the learning rate of the I-th layer.}  \]
    
    \citeauthor{howard2018universal} introduced slanted triangular learning rate (STLR) to make the model parameters adapt to the task-specific text features. In this stage, the learning rate first increases linearly and then decays linearly.   
    \item \textbf{Target task classifier fine-tuning:}
    In the final stage of ULMFiT, the model is trained with two extra linear blocks. Each block uses batch normalization and dropout. The intermediate layer is activated with ReLU, while the final linear layer is activated with Softmax. The most important aspect of the transfer learning approach is fine-tuning the target classifier. Hence, Gradual unfreezing is used to fine-tune rather than training all layers at once, leading to catastrophic forgetting.

\end{enumerate}
\subsubsection{MuRIL}
MuRIL or Multilingual Representation for Indian Languages \citep{khanuja2021muril} was introduced to boost Indian Natural Language Understanding in 17 Indian languages. Introduced by Google Research India, it uses BERT \citep{devlin-etal-2019-bert, hegde2021uvce} architecture but differs from mBERT as it is trained on translation and transliteration pairs. It uses the IndicNLP-Transliteration library to transliterate the Wikipedia dataset and also the Dakshina \footnote{https://github.com/google-research-datasets/dakshina} dataset to train the model on transliterated data. This approach promises better accuracy on transliterated data. MuRIL trained it with a Masked Language Modeling (MLM) to give a maximum of 80 predictions with 4096 as the batch size and a maximum sequence length of 512.
 
\subsection{XLM-RoBERTa}
XLM-Roberta \citep{conneau2020unsupervised}, a model by the Facebook AI team, is a transformer-based Masked Language Model (MLM) from over 100 languages, including the low resourced languages released as an update to the previous XLM-100 model \citep{lample2019crosslingual}. It follows the training routine as the RoBERTa model \citep{liu2019roberta} which adds it to its name. It distinguishes itself from the other models with its more training data up to 2.5TB of the new at the time of release CommonCrawl data\footnote{https://commoncrawl.org/} and more languages. The model discusses various topics like the constraints of the multilingual MLMs and that the multilingual model provides better results than monolingual models with state-of-the-art outcomes for cross-lingual classification, question answering, and various other tasks. We have used \emph{xlm-roberta-large} from HuggingFace\footnote{https://huggingface.co/} to predict the pseudo labels for the transliterated data due to its robust results and the fact that it was the largest multilingual language model available at the time of this paper.
\begin{equation}\label{eqn3}
    FFN(x)=max(0,x * W_1 + b_1) W_2 + b_2
\end{equation}
\begin{figure}
    \centering
    \includegraphics[width=\tblwidth, height=8cm]{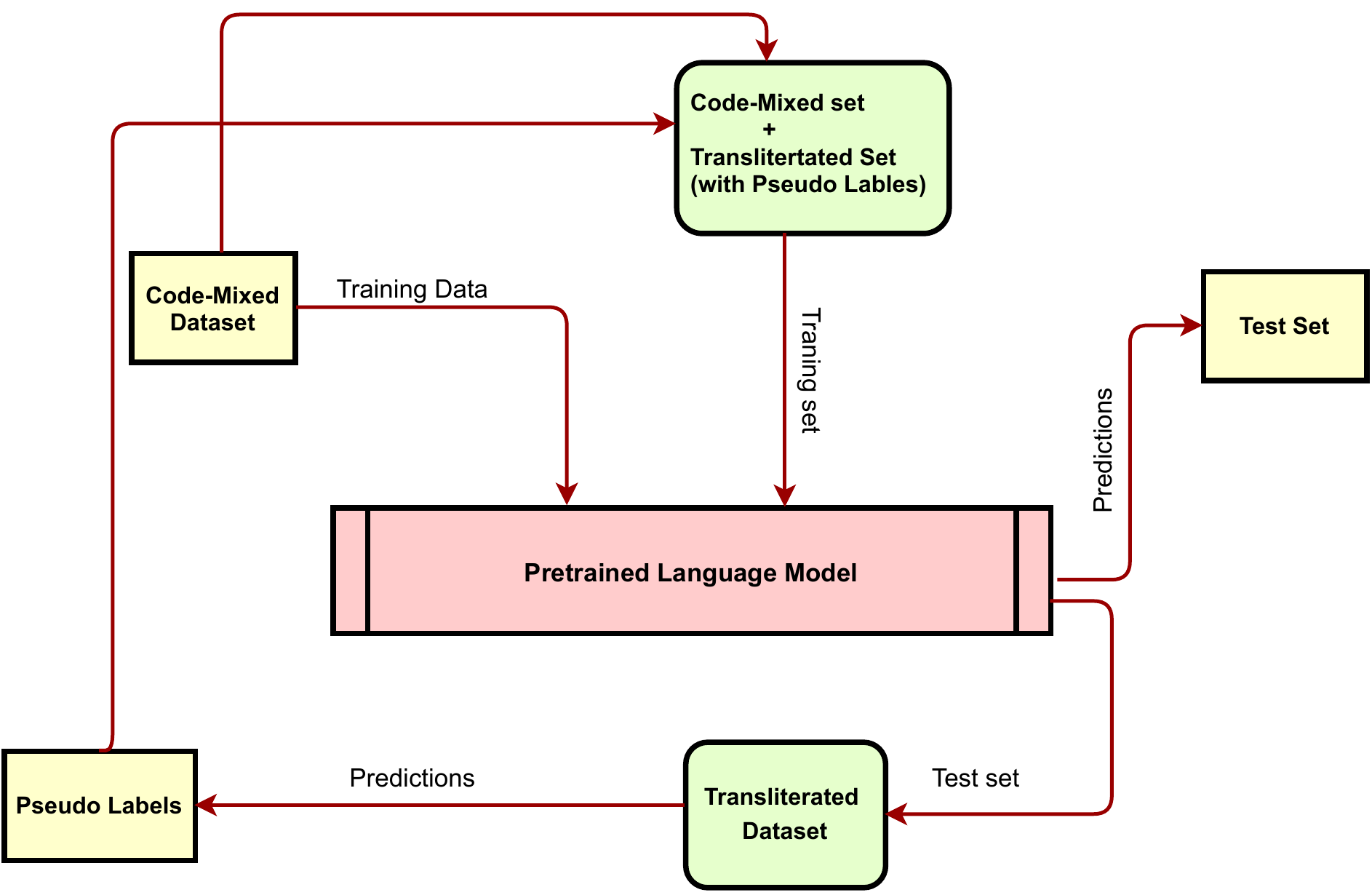}
    \caption{Generating Pseudo Labels on the transliterated dataset.}
    \label{pseudo_labeling}
\end{figure}
\RestyleAlgo{ruled}

\SetKwComment{Comment}{/* }{ */}
\SetKwInOut{Transliteration}{Transliteration}
\SetKwInOut{Parameter}{parameter}
\SetKwInOut{Return}{Return}
\SetKwInOut{Loss}{Loss Function}
\SetKwInOut{Dataset}{Dataset}
\begin{algorithm}
\caption{Proposed Approach}\label{algo_disjdecomp}
\SetKwData{Left}{left}
\SetKwData{This}{this}
\SetKwData{Up}{up}
\SetKwFunction{Union}{Union}
\SetKwFunction{FindCompress}{FindCompress}
\SetKwInOut{Input}{Input}
\SetKwInOut{Output}{Output}
\Input{Code-mixed dataset \textit{CM}}
\Output{Class Label L, generated for the transliterated dataset using Pseudo-labeling}
\emph{Let $C = c_1, c_2, .., c_n$ where $C_i \in (kn-en/ml-en/ta-en)$  be the code-mixed sentences} \\ 
\Transliteration{Code-mixed dataset CM to Kannada/Malayalam/Tamil}
\emph{Let $T = t_1, t_2, ..., t_n$ where $T_i \in (kn/ml/ta)$}\\
\For{$T_i \in T$}{
\emph{Extract cross-lingual embeddings $E[T_i]$ from XLM-RoBERTa-large, where \textit{$E_{size}[T_i]=1024$}} \\
$Q = E[T_i] * W^Q$\\
$K = E[T_i] * W^K$\\
$V = E[T_i] * W^V$\\
where $W^Q, W^K, W^V$ are the weight projections for Q, K, and V\\
\For{$i=0$ \KwTo $16$}{\tcp*[r]{num(heads) = 16}
\textit{$Attention(Q_i, K_i, V_i) = softmax(\frac{QK^T}{\sqrt{d_{k}}})V$}\\ }
$MultiHead(Q,K,V)=Concat(Head_1,Head_2,....,head_{16})W_O$\\
    where, $Head_i=Attention(QW_i^Q,KW_i^K,VW_i^V)$
}
\Loss{$L_{CE} = \sum_{i}^C t_{i}\log (s_{i}) $}\tcp*[r]{{\(t_{i}\) and \(s_{i}\) are the ground truths for every class i in C}}
\emph{Obtain the embeddings from the pooler output, $T_E$ and feed it as input to an LSTM block}\\
\For{$j = 1$ \KwTo $D$}{
\tcp*[r] {D = Number of memory blocks in LSTM}
$I_t = \sigma(w_i.[h_{t-1}, T_E] + b_i)$ \tcp*[f] {$I_t$ = Input Gate}\\

$F_t = \sigma(w_f.[h_{t-1}, T_E] + b_f)$  \tcp*[f] {$F_t$ = Forget Gate}\\

$O_t= \sigma(w_o.[h_{t-1}, T_E] + b_o)$  \tcp*[f] {$O_t$ = Output Gate}\\

$C_i = \tanh(w_c.[h_{t-1}, T_E] + b_c)$  \tcp*[f] {$C_i$ = Updating the memory cell, $C_i$}\\}
\emph{Obtain the final output of the LSTM Block}\\
\emph{$T = t_1, t_2, ..., t_n$ where $T_i \in (kn/ml/ta)$ is used as the test set}\\
\emph{Output $O$ is fed into a softmax layer to obtain probabilities (P) of all classes} \tcp*[f] {$N(Class) = 6, 6, 5 (kn, ta, ml)$}\\
$Softmax(z) = \frac{e^{z_i}}{\sum_{j=1}^K e^{z_j}} $ \tcp*[f] {softmax over \textbf{K} classes}\\
\emph{Output Class, L = argmax(P)} \tcp*[f]{Obtaining Pseudo-labels on the transliterated test set}\\
\Return {Label $L$ (Pseudo-labels)}
\emph{Combining $T = t_1, t_2, ..., t_n$ and $C = c_1, c_2, .., c_n$ into a single dataset }\\
\Dataset{CM-TRA dataset}
\end{algorithm}
\subsection{Transliteration}
We transliterate the dataset with the help of the AI4Bharat transliteration application, Indic Deep-Xlit Engine\footnote{https://github.com/AI4Bharat/IndianNLP-Transliteration}. This deep transliteration engine can translate from Roman script to significant Indian languages and contain under-represented and resourced languages. Its architecture features a seq2seq model with RNNs (Recurrent neural network) and encoder-decoders. The model efficiently learns all the embeddings and weights, and the decoder acquires \emph{topk} predictions which are ranked, and the most likely words are predicted \citep{bahdanau2016neural}. The training datasets were transliterated for Malayalam, Tamil and Kannada. Various experiments were conducted using the transliterated dataset. Firstly, the models featured in \cite{yasaswini-etal-2021-iiitt} were fed with the transliterated data to compare the F1 scores obtained. Later, the transliterated dataset was used to perform pseudo labelling. 

\subsection{Pseudo Labeling}
While the machine learning algorithms and models to produce state-of-the-art results evolve, the limitation of datasets and the uneven imbalance in the classes still poses a problem to researchers. The field of image processing deals with this problem with classical data augmentation techniques like rotation, cropping, zooming and new modus operandi, including GANs and Style transfer \citep{8388338}. Pseudo labeling is one technique that can be used to handle this problem in Natural Language Processing. Pseudo-labeling is a semi-supervised learning technique in which a model is trained on a set of labeled data and used to predict the unlabeled test data. The test data, along with their supposed labels, are merged with the training data for additional training \citep{aroyehun-gelbukh-2018-aggression}. We have implemented this technique slightly different. The test data has been left untouched, and the labels have been predicted for the transliterated training dataset. Furthermore, the transliterated and the unprocessed training dataset have been collectively used to train the models. 

\section{Results}
\label{results}
\begin{figure}
    \centering
    \includegraphics[width=12cm, height=8cm,angle=90]{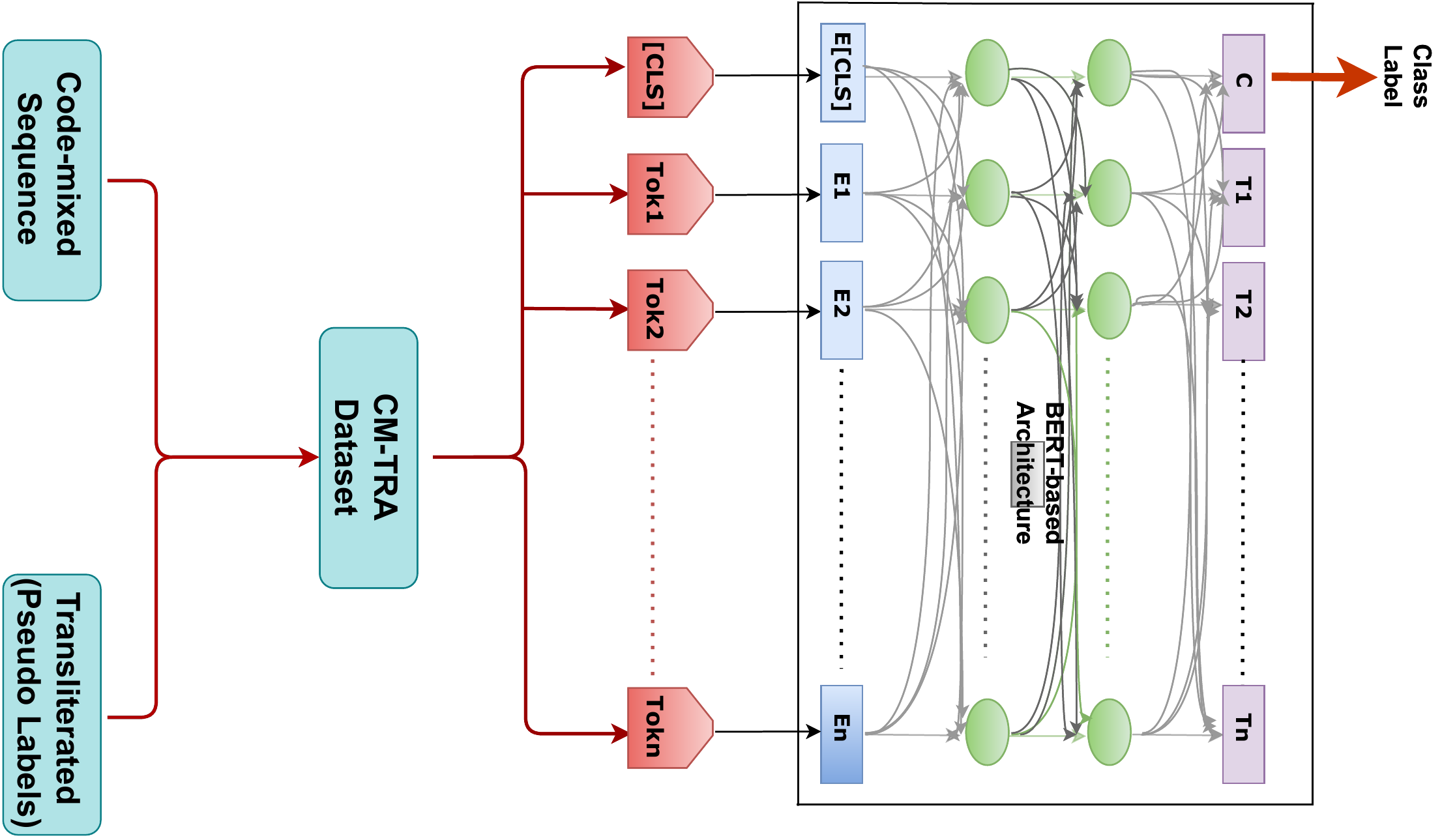}
    \caption{Fine-tuning BERT-based models on the CM-TRA dataset.}
    \label{fig:cmtra-bert}
\end{figure}
\begin{table}[width=.9\linewidth,cols=4,pos=h]
\caption{Weighted Precision, Weighted Recall, and Weighted F1-scores of offensive language detection models on the three datasets}\label{tbl4}
\begin{tabular*}{\tblwidth}{@{} L|RRR|RRR|RRR@{} }
\toprule
\textbf{Model} &\multicolumn{9}{c}{\textbf{Code-Mixed Dataset}} \\
\midrule
& \multicolumn{3}{c|}{Malayalam} & \multicolumn{3}{c|}{Tamil} & \multicolumn{3}{c}{Kannada}\\ \midrule
&\textbf{P} & \textbf{R} & \textbf{F1} & \textbf{P} & \textbf{R} & \textbf{F1} &\textbf{P} & \textbf{R} & \textbf{F1} \\  \midrule
mBERT  &0.9195 &0.9410 & 0.9301 & 0.7461 & 0.7664 & 0.7556 & 0.6863 & 0.7082& 0.6936 \\
XLM-R & 0.9206 & 0.9380 & 0.9288 &0.5275 &0.7263 &0.6112 &0.6449 & 0.7326 & 0.6851\\ 
DistilmBERT  & 0.9411 & 0.9520 & 0.9465 &0.7368 &0.7632 &0.7489 &0.6789 & 0.7249& 0.7010\\
MURiL &0.7780 &0.8821 &0.8268 & 0.5275 & 0.7263 & 0.6112 &0.3012 &0.5488 &0.3890 \\
IndicBERT & 0.9572 & 0.9600 & 0.9568 & 0.7150 & 0.7454 & 0.7287 & 0.6714 & 0.6992 & 0.6809\\
ULMFiT &0.9643 & 0.9580 & 0.9603 & 0.8220 & 0.7650 & 0.7895 & 0.7186 & 0.6864 & 0.7000\\\midrule
& \multicolumn{9}{c}{\textbf{Transliterated Dataset}} \\ \midrule
& \multicolumn{3}{c|}{Malayalam} & \multicolumn{3}{c|}{Tamil} & \multicolumn{3}{c}{Kannada}\\ \midrule
&\textbf{P} & \textbf{R} & \textbf{F1} & \textbf{P} & \textbf{R} & \textbf{F1} &\textbf{P} & \textbf{R} & \textbf{F1} \\  \midrule
mBERT&0.9023&0.9398&0.9202&0.7063&0.7648&0.7286&0.6779&0.7389&0.7002 \\
XLM-R &0.8902&0.9265&0.9080 &0.5538&0.7320&0.6290&0.6369&0.7198&0.6750\\
DistilmBERT&0.9089&0.9370&0.9199&0.7248&0.7571&0.7390&0.6789&0.7249&0.7010 \\  
MuRIL &0.9039&0.9405&0.9218&0.5275 &0.7263 &0.6112 & 0.6432&0.7249&0.6815\\
IndicBERT &0.9305&0.9445&0.9373&0.7194&0.7354&0.7263&0.6433&0.6722&0.6558\\
ULMFiT & 0.9521 & 0.9505 & 0.9508 & 0.8033 & 0.7682 & 0.7842 &0.7304 & 0.6979 & 0.7115\\
\midrule
& \multicolumn{9}{c}{\textbf{CM-TRA}}\\ \midrule
& \multicolumn{3}{c|}{Malayalam} & \multicolumn{3}{c|}{Tamil} & \multicolumn{3}{c}{Kannada}\\ \midrule
&\textbf{P} & \textbf{R} & \textbf{F1} & \textbf{P} & \textbf{R} & \textbf{F1} &\textbf{P} & \textbf{R} & \textbf{F1} \\  \midrule
mBERT&0.9468&0.9535&0.9478 &0.6865&0.7432&0.7026 &0.6048&0.6517&0.6188 \\
XLM-R &0.9370&0.9410&0.9366 &0.7284&0.7609&0.7427& 0.6997&0.7455&0.7029  \\
DistilmBERT&  0.9582&0.9575&0.9537& 0.7414&0.7516&0.7461 &0.7008&0.7198&0.7037 \\  
MuRIL & 0.7780&0.8821&0.8268&0.7081&0.7511&0.7045 & 0.6407&0.7249&0.6801 \\
IndicBERT &0.9306&0.9465 &0.9380&0.6867&0.7516&0.7057& 0.5937&0.6671&0.6235\\
ULMFiT &\textbf{0.9649}&\textbf{0.9610}&\textbf{0.9624}&\textbf{0.8203}&\textbf{0.7719}&\textbf{0.7934}&\textbf{0.7576}&\textbf{0.7104}&\textbf{0.7306}\\
\bottomrule
\end{tabular*}
\end{table}
In this section, we report the precision (P), recall (R), F1 scores (F1) of our transformer-based models to identify offensive comments/posts and further classify them offensive-targeted-insult-individual, offensive-targeted-insult-group, offensive-targeted-insult-other, offensive untargeted and not-in-intended-language. The TP, FP, FN, TN values of a class are defined as follows -
\begin{enumerate}
    \item \emph{TP (True Positive)} - the value of the actual class is positive, and the value of the predicted class is also positive.
    \item \emph{FP (False Positive)} - the value of the actual class is negative, but the value of the predicted class is positive.
    \item \emph{FN (False Negative)} - the value of the actual class is positive, but the value of the predicted class is negative.
    \item \emph{TN (True  Negative)} - the value of the actual class is negative, and the value of the predicted class is negative.
\end{enumerate}

\begin{equation}
\emph{Precision(P) = \(\frac{TP}{TP+FP}\)}
\end{equation}

\begin{equation}
\emph{Recall(R) = \(\frac{TP}{TP+FN}\)}
\end{equation}

\begin{equation}
\emph{F1 score(F1) = \(\frac{2 \mbox{*}R \mbox{*}P}{R+P}\)}
\end{equation}

\begin{equation}
P_{\text {weighted }}=\sum_{i=1}^{L}(P \text { of } i \times \text { Weight of } i)
\end{equation}

\begin{equation}
R_{\text {weighted }}=\sum_{i=1}^{L}(R \text { of } i \times \text { Weight of } i)
\end{equation}

\begin{equation}
F 1_{\text {weighted }}=\sum_{i=1}^{L}(F 1 \text { of } i \times \text { Weight of } i)
\end{equation}

For future reference, we refer to the dataset comprising the custom code-mixed and transliterated set as CM-TRA. Table \ref{tbl4} shows the weighted average F1-scores of various transformer-based models trained and evaluated on code-mixed, transliterated and CM-TRA datasets of Malayalam, Tamil, and Kannada. 

XLM-Roberta Large \citep{conneau2020unsupervised}, the largest multilingual model available at the time of our work, was used to predict the labels for the transliterated data. Later, we used the combined data to train various top-performing models. We expected this approach to produce a considerable increase in the F1 scores, as shown in Figure \ref{pseudo_labeling}. We have discussed our proposed approach in Algorithm \ref{algo_disjdecomp}.

However, test set we are using is the holdout set of \citep{chakravarthi2021dravidiancodemix}, we do not employ any cross-validation techniques in our experiments. As observed in Table \ref{tbl2} and Table \ref{tbl3} representing the class-wise distribution of the training and test sets. To address the issue of class imbalance, we use class weighting. The inverse the weights of each class and pass it as tensor while computing the loss during training \citep{hande2021benchmarking}. While this approach had improved the recall of the classes with low samples, it drastically decreases the performance of the large sampled classes, effectively reducing the overall weighted F1-Scores. Hence, we refrain from using class weighting for computing the loss during training, and revert back to the traditional computation of loss where all classes are treated with equal importance.

We have experimented with models like multilingual BERT, XLM-RoBERTa, DistilmBERT, MURiL, IndicBERT and ULMFiT. We observe that the ULMFiT model is the better performing model on code-mixed datasets of Malayalam, Tamil with F1-scores 0.9603, 0.7895 respectively, and DistilmBERT is the better performing model on Kannada of F1-score 0.7000. MURiL is the model that gave poor results on Tamil, Malayalam and Kannada code-mixed datasets with F1-scores of 0.6112, 0.8268 and 0.3890, respectively, compared with other models. One of the reasons for poor performance could be class imbalances and code-mixed and writing in non-native languages. The remaining models gave a relatively good performance on the three language code-mixed datasets. The process of fine-tuning on the custom dataset CM-TRA has been computed as shown in Figure \ref{fig:cmtra-bert}.
 \begin{table}[width=.9\linewidth,cols=4,pos=htbp]
     \centering
     \begin{tabular}{l|rrrr}
     \toprule
       Approach   &  Precision & Recall & F1-Score & Rank \\
       \midrule
       \textbf{CM-TRA-ULMFiT} & \textbf{0.82} & 0.77 & \textbf{0.79} & \textbf{1}\\
        \citep{saha-etal-2021-hate} & 0.78&  0.78& 0.78& 2\\
        \citep{kedia-nandy-2021-indicnlp}& 0.75& \textbf{0.79}& 0.77&3\\
       \citep{zhao-tao-2021-zyj123} &0.75& 0.77& 0.76&4\\ 
        \citep{jayanthi-gupta-2021-sj}& 0.75& \textbf{0.79} &0.76&4\\ 
       \citep{sharif-etal-2021-nlp}& 0.75&0.78&0.76&4\\
       \citep{li-2021-codewithzichao}& 0.74 &0.77& 0.75&7\\
        \citep{huang-bai-2021-hub}&0.73& 0.78& 0.75&7\\
        \citep{balouchzahi-etal-2021-mucs}&  0.74& 0.77 &0.75&7\\
        \citep{tula-etal-2021-bitions} &0.74& 0.77 &0.75&7\\
       \citep{ghanghor-etal-2021-iiitk}& 0.74 &0.77& 0.75&7\\
        \citep{dowlagar-mamidi-2021-graph}& 0.74 &0.75 &0.75&7\\
        \citep{chen-kong-2021-cs}& 0.74& 0.75& 0.74 &13\\
       \citep{vasantharajan-thayasivam-2021-hypers} &0.71 &0.76& 0.73& 14\\
       \citep{b-a-2021-ssncse}& 0.74& 0.73& 0.73& 14\\
        \citep{garain-etal-2021-junlp}&0.71& 0.74& 0.72&16\\ 
        \citep{yasaswini-etal-2021-iiitt}& 0.70& 0.73& 0.71&17 \\
        \citep{dave-etal-2021-irnlp}& 0.72 &0.77& 0.71&17 \\
        \citep{renjit-idicula-2021-cusatnlp} &0.67 &0.71& 0.69& 19\\
        \citep{k-etal-2021-amrita}& 0.64& 0.62 &0.62&20\\
        \citep{andrew-2021-judithjeyafreedaandrew} &0.54 &0.73& 0.61 &21\\
       \bottomrule
     \end{tabular}
     \caption{Comparisons of the existing models that were developed for the Tamil dataset with other existing models on the dataset. Ranks are based on the descending order of the weighted F1-Scores.}
     \label{tab:comp_tam}
 \end{table}
 \begin{figure}[htbp]
    \begin{center}
     \includegraphics[width=0.5\linewidth]{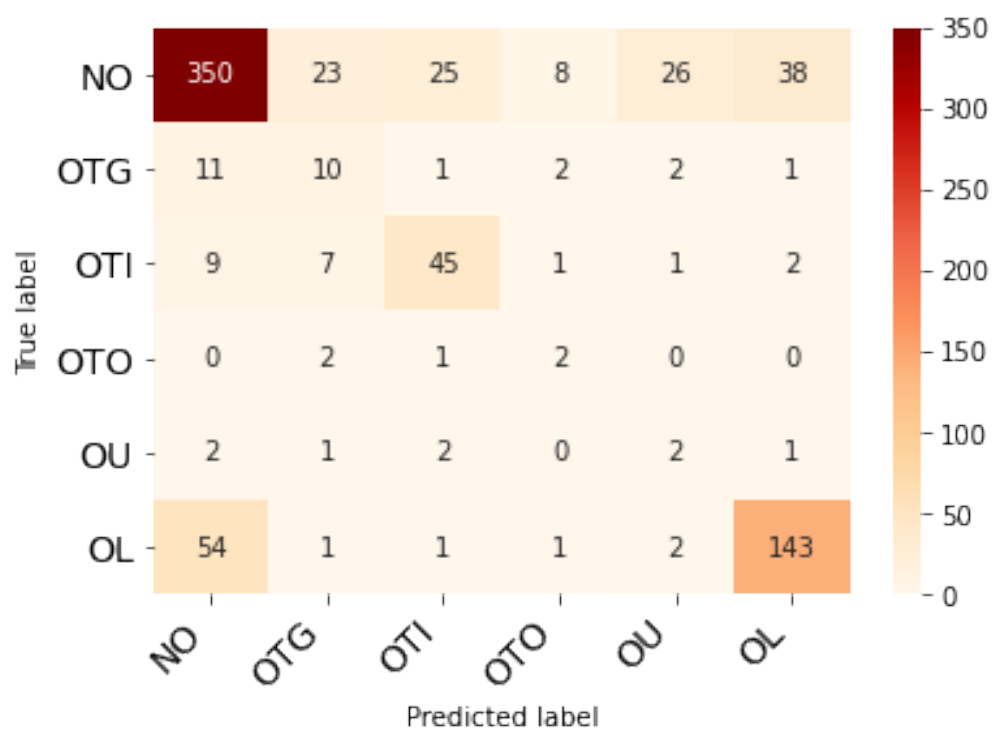}
     \caption{Heatmap of the confusion matrix for the test set of code-mixed English-Kannada} 
     \label{fig:kan}
         \end{center}
\end{figure} 
 \begin{table}[width=.9\linewidth,cols=4,pos=htbp]
     \centering
     \begin{tabular}{l|rrrr}
     \toprule
       Approach   &  Precision & Recall & F1-Score & Rank \\
       \midrule
       \citep{jayanthi-gupta-2021-sj}   & 0.73 & \textbf{0.78} &\textbf{0.75} &1\\
       \citep{saha-etal-2021-hate} & \textbf{0.76}&0.76& 0.74&2\\
      \textbf{CM-TRA-ULMFiT} & \textbf{0.76} & 0.71 & 0.73 &\textbf{3}\\
       \citep{kedia-nandy-2021-indicnlp} & 0.71&0.74& 0.72&4\\
       \citep{li-2021-codewithzichao} & 0.70& 0.75& 0.72&4\\
      \citep{ghanghor-etal-2021-iiitk} & 0.70& 0.75& 0.72&4\\
       \citep{sharif-etal-2021-nlp} & 0.70& 0.74 &0.71&7\\
       \citep{vasantharajan-thayasivam-2021-hypers}&0.69 &0.72& 0.70&8\\
       \citep{tula-etal-2021-bitions}&0.69 &0.72& 0.70&8\\
       \citep{b-a-2021-ssncse}&0.71& 0.74& 0.70&8\\
       \citep{balouchzahi-etal-2021-mucs} & 0.68 & 0.72 & 0.69&11\\
        \citep{zhao-tao-2021-zyj123} & 0.65 & 0.74  &0.69&11\\
        \citep{garain-etal-2021-junlp}& 0.62 & 0.71 & 0.66&13\\
        \citep{dowlagar-mamidi-2021-graph} & 0.66 & 0.65 & 0.65&14\\
        \citep{chen-kong-2021-cs-dravidianlangtech}  &0.64  &0.67 & 0.64&15\\
        \citep{huang-bai-2021-hub} & 0.65 & 0.69  &0.64&15\\ 
        \citep{dave-etal-2021-irnlp} &0.69 & 0.69 & 0.64&15\\
        \citep{andrew-2021-judithjeyafreedaandrew}  &0.66  &0.67  &0.63&18\\
        \citep{renjit-idicula-2021-cusatnlp} &  0.62  &0.63  &0.62&19 \\
        \citep{k-etal-2021-amrita}  &0.65  &0.54 & 0.58&20 \\ 
        \citep{yasaswini-etal-2021-iiitt}  &0.46 & 0.48 & 0.47&21 \\
        \citep{que-2021-simon}  &0.60 & 0.30 & 0.33 & 22\\
       \bottomrule
     \end{tabular}
     \caption{Comparisons of the existing models developed for the code-mixed Kannada dataset.}
     \label{tab:comp_kan}
 \end{table}
 \begin{table}[width=0.9\linewidth, cols=4, pos=htbp]
     \centering
     \begin{tabular}{l|rrrr}
     \toprule
      Approach   &  Precision & Recall & F1-Score & Rank\\
      \midrule
        \citep{saha-etal-2021-hate}& \textbf{0.97}& \textbf{0.97}& \textbf{0.97}&1 \\
        \citep{balouchzahi-etal-2021-mucs} &0.97 &0.97 &0.97&1\\
        \citep{kedia-nandy-2021-indicnlp} &0.97 &0.97& 0.97 &1\\
        \citep{tula-etal-2021-bitions}& 0.97 &0.97& 0.97 &1\\
        \textbf{CM-TRA-ULMFIT} &  {0.96}& {0.96}& {0.96}&\textbf{5}\\
        \citep{vasantharajan-thayasivam-2021-hypers}& 0.96& 0.96 &0.96& 6\\
        \citep{dowlagar-mamidi-2021-graph}& 0.96 &0.96& 0.96&6\\
        \citep{jayanthi-gupta-2021-sj}&0.97 &0.97 &0.96&6\\
        \citep{renjit-idicula-2021-cusatnlp}& 0.95 &0.95 &0.95& 9\\
        \citep{ghanghor-etal-2021-iiitk}& 0.94& 0.95& 0.95&9\\
        \citep{b-a-2021-ssncse} & 0.95& 0.96 &0.95& 9\\
        \citep{dave-etal-2021-irnlp} &0.96 &0.96 &0.95 &9\\
        \citep{li-2021-codewithzichao}& 0.93 &0.94 &0.94&13\\
        \citep{andrew-2021-judithjeyafreedaandrew}& 0.94 &0.94& 0.93 &13\\
        \citep{chen-kong-2021-cs-dravidianlangtech} &0.92 &0.94 &0.93 &13\\
       \citep{yang-2021-maoqin}  & 0.91&0.94 &0.93 &13\\
        \citep{sharif-etal-2021-nlp} &0.92& 0.94 &0.93 &13\\
        \citep{zhao-tao-2021-zyj123}& 0.91 &0.94& 0.92 &18\\
       \citep{huang-bai-2021-hub}& 0.89& 0.93 &0.91&19\\ 
        \citep{yasaswini-etal-2021-iiitt}& 0.84& 0.87& 0.86&20\\
        \citep{k-etal-2021-amrita} &0.90& 0.82& 0.85& 21\\
        \citep{nair-fernandes-2021-professionals}& 0.89& 0.84 &0.85 &21\\
        \citep{garain-etal-2021-junlp}& 0.77 &0.43& 0.54 &23\\
\bottomrule
     \end{tabular}
     \caption{Comparisons of the existing models developed for the code-mixed Malayalam dataset.}
     \label{tab:comp_mal}
 \end{table}

There is a significant improvement in the MURiL model on Malayalam and Kannada transliterated datasets of F1-scores 0.9218 and 0.6815, respectively, while there is no improvement of the model on the Tamil transliterated dataset. The other models showed a marginal decrement in the performance on Malayalam transliterated dataset. In the Tamil transliterated dataset, except the XLM-RoBERTa model, the remaining models performed relatively poor performance, and the XLM-RoBERTa model showed an improvement with an F1-score 0.6290. The multilingual BERT model and ULMFiT model showed a slight increase in the model performance on Kannada transliterated. Surprisingly, on transliterated datasets, ULMFiT still gave better results when compared to the other models. The exceptional fine-tuning methods of the ULMFiT model may also result in giving better performance over other models.     

To our expectations, the performance of the models is noticed to be ameliorated on CM-TRA Tamil, Malayalam and Kannada datasets. In comparison to the performance of the models on transliterated datasets, DistilmBERT and XLM-RoBERTa models showed a significant enhancement of the performance on the three languages. It is observed that the IndicBERT model showed no improvement on the datasets. Overall, the ULMFiT model gave promising results on CM-TRA datasets of all the three languages with F1-scores of 0.9624 (Malayalam), 0.7934 (Tamil) and 0.7306 (Kannada).

\begin{table}[width=0.9\linewidth, cols=4, pos=htbp]
\begin{tabular}{|l|rrrr|rrrr|}
\toprule
                              & \multicolumn{4}{c}{Kannada}                                                                            & \multicolumn{4}{c|}{Malayalam}                                                                         \\ \midrule
                              & \multicolumn{1}{r|}{Precision} & \multicolumn{1}{r|}{Recall} & \multicolumn{1}{r|}{F1-Score} & Support & \multicolumn{1}{r|}{Precision} & \multicolumn{1}{r|}{Recall} & \multicolumn{1}{r|}{F1-Score} & Support \\ \midrule
Not Offensive                 & 0.8216                         & 0.7447                      & 0.7812                        & 417     & 0.9875                         & 0.9710                      & 0.9792                        & 1,765   \\  
Other Language                & 0.7730                         & 0.7079                      & 0.7390                        & 185     & 0.8408                         & 0.9496                      & 0.8919                        & 157     \\  
Offensive Targeted Individual & 0.6000                         & 0.6923                      & 0.6429                        & 75      & 0.5926                         & 0.8421                      & 0.6957                        & 27      \\ 
Offensive Targeted Group      & 0.2273                         & 0.3704                      & 0.2817                        & 44      & 0.5217                         & 0.6667                      & 0.5854                        & 23      \\  
Offensive Targeted Others     & 0.1429                         & 0.4000                      & 0.2105                        & 14      & -                              & -                           & -                             & -       \\ 
Offensive Untargeted          & 0.0606                         & 0.2500                      & 0.0976                        & 33      & 0.6897                         & 0.6667                      & 0.6667                        & 29      \\ \midrule
Accuracy                      &                                &                             & 0.7104                        & 768     &                                &                             & 0.9610                        & 2,001   \\  
Macro Average                 & 0.4376                         & 0.5275                      & 0.4588                        & 768     & 0.7265                         & 0.8192                      & 0.7660                        & 2,001   \\ 
Weighted Average              & \multicolumn{1}{r}{0.7576}    & \multicolumn{1}{r}{0.7104} & \multicolumn{1}{r}{0.7306}   & 768     & \multicolumn{1}{r}{0.9649}    & \multicolumn{1}{r}{0.9610} & \multicolumn{1}{r}{0.9624}   & 2,001   \\ \bottomrule
\end{tabular}
\caption{Classification report of CM-TRA-ULMFiT on the Kannada and Malayalam test sets.}
\label{cm_kan_mal}
\end{table}
\begin{table}[width=0.9\linewidth, cols=4, pos=htbp]
\begin{tabular}{|l|rrrr|}
\toprule
                              & \multicolumn{4}{c|}{Tamil}  \\ \midrule
                              & \multicolumn{1}{r|}{Precision} & \multicolumn{1}{r|}{Recall} & \multicolumn{1}{r|}{F1-Score} & Support \\ \midrule
Not Offensive                 & 0.9285                         & 0.8456                      & 0.8851                        & 3,190  \\  
Other Language                & 0.8000                         & 0.8421                      & 0.8205                        & 165     \\ 
Offensive Targeted Individual & 0.3492                         & 0.4527                      & 0.3943                        & 315     \\  
Offensive Targeted Group      & 0.2639                         & 0.3689                      & 0.3077                        & 288    \\ 
Offensive Targeted Others     & 0.0                            & 0.0                         & 0.0                           & 71     \\  
Offensive Untargeted          & 0.3098                         & 0.3958                      & 0.3476                        & 368     \\ \midrule
Accuracy                      &                                &                             & 0.7719                        & 4,392   \\ 
Macro Average                 & 0.4419                         & 0.4842                      & 0.4592                        & 4,392 \\ 
Weighted Average              & \multicolumn{1}{r}{0.8203}    & \multicolumn{1}{r}{0.7719} & \multicolumn{1}{r}{0.7934}   & 4,392    \\ \bottomrule
\end{tabular}
\caption{Classification report of CM-TRA-ULMFiT on the Tamil test sets.}\label{cr_tam}
\end{table}

One of the essential things we can observe in the results is the slight decrease in the F1 score for the transliterated data compared to the code-mixed dataset. This drop can be because all the pretrained models are primarily trained on large English datasets and fewer sentences belonging to a particular language. So, when fine-tuned on only the Dravidian language, the model might not detect the other languages in the test dataset well. Our code-mixed dataset shows the predominant use of English with the Dravidian languages. The presence of English sentences in the code-mixed data seems to be the determining point for this decrease. The other factor responsible for this can be the inconsistency of transliterating the data using the Ai4Bharat transliteration application. The pseudo labels obtained from these data may not be accurate, causing the models to get misled when trained on transliterated data.
\begin{figure}[htbp]
    \begin{center}
     \includegraphics[width=0.5\linewidth]{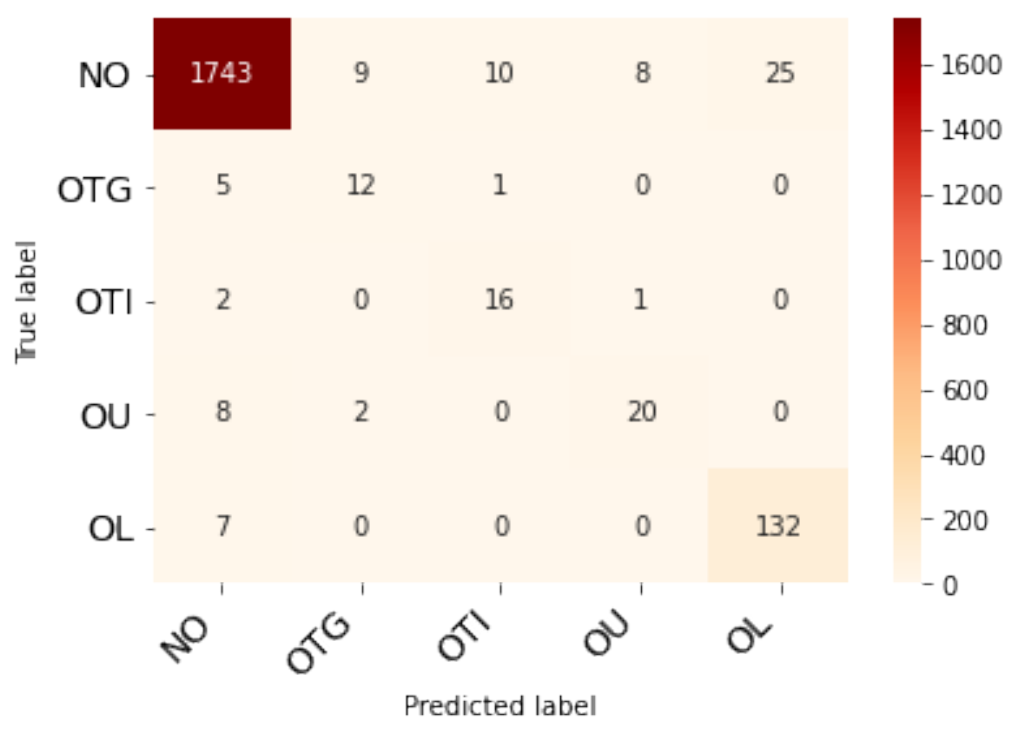}
     \caption{Heatmap of the confusion matrix for the test set of code-mixed Malayalam-English} 
     \label{fig:mal}
         \end{center}
\end{figure} 

\begin{figure}[htpb]
    \begin{center}
     \includegraphics[width=0.5\linewidth]{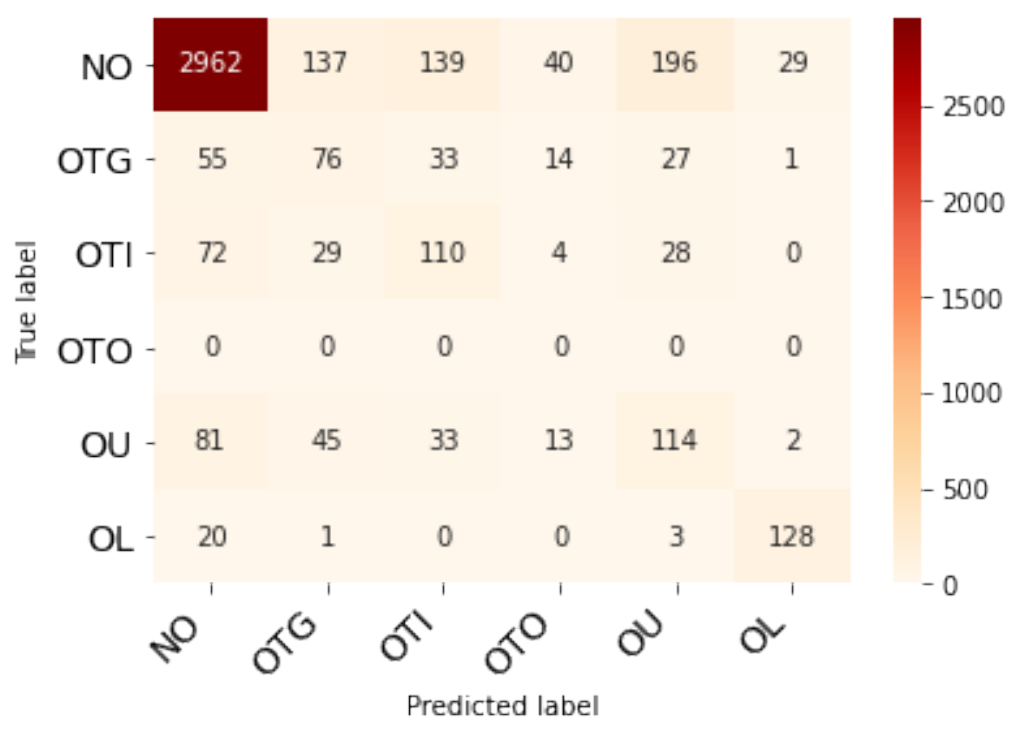}
     \caption{Heatmap of the confusion matrix for the test set of code-mixed Tamil-English} 
     \label{fig:tam}
         \end{center}
\end{figure}
However, as expected, the brainchild of our research, the models, when trained on the CM-TRA dataset, gave exceptional results. With the combination of the code-mixed and transliterated data, the model successfully learned the code-mixed and language aspects. One of the trends observed is that the BERT based models, including mBERT, DistilmBERT, IndicBERT, show declining F1 scores for the CM-TRA dataset, especially for the languages of Kannada and Tamil. This decline could indicate the types of Tamil and Kannada datasets used during pretraining for the models compared to the code-mixed data scraped from the internet. IndicBERT, which is extensively pretrained in Indian languages, shows a rapid decline in the results. XLM-RoBERTa, though it possesses BERT architecture, is trained on transliterated datasets, which probably is the reason for not showing this decreasing trend. However, due to superior pretraining strategies, ULMFiT fared well on the dataset and procured the best F1 scores for all three languages on the CM-TRA dataset. The classification reports for the best performing models are tabulated in Table \ref{cm_kan_mal} and Table \ref{cr_tam}.
\begin{figure}[htpb]
    \centering
    \includegraphics[width=0.8\linewidth, height=12cm]{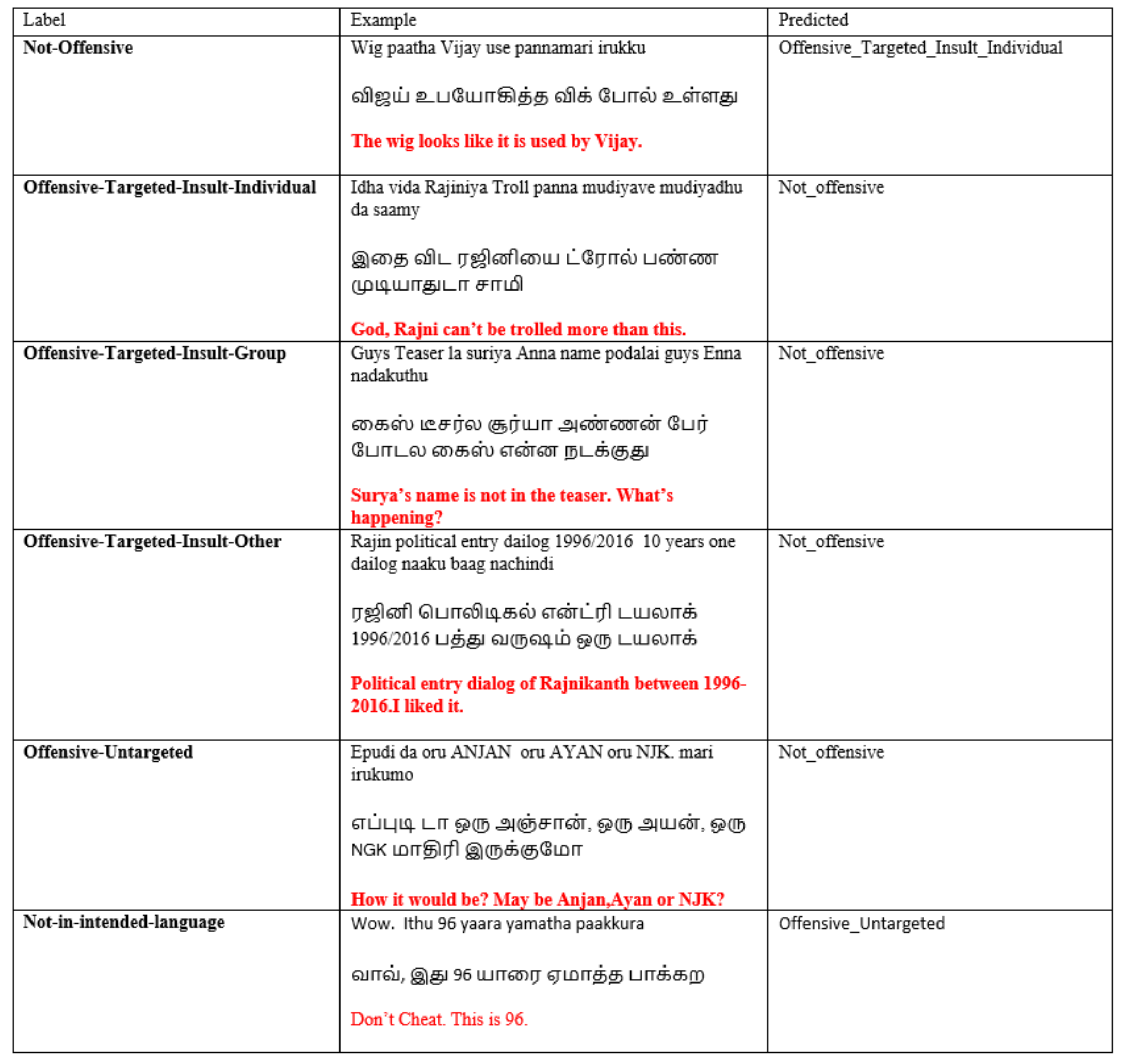}
    \caption{Wrong Predictions in the Tamil dataset}
    \label{fig:tam_error}
\end{figure}

\begin{figure}[htpb]
    \centering
    \includegraphics[width=0.8\linewidth, height=12cm]{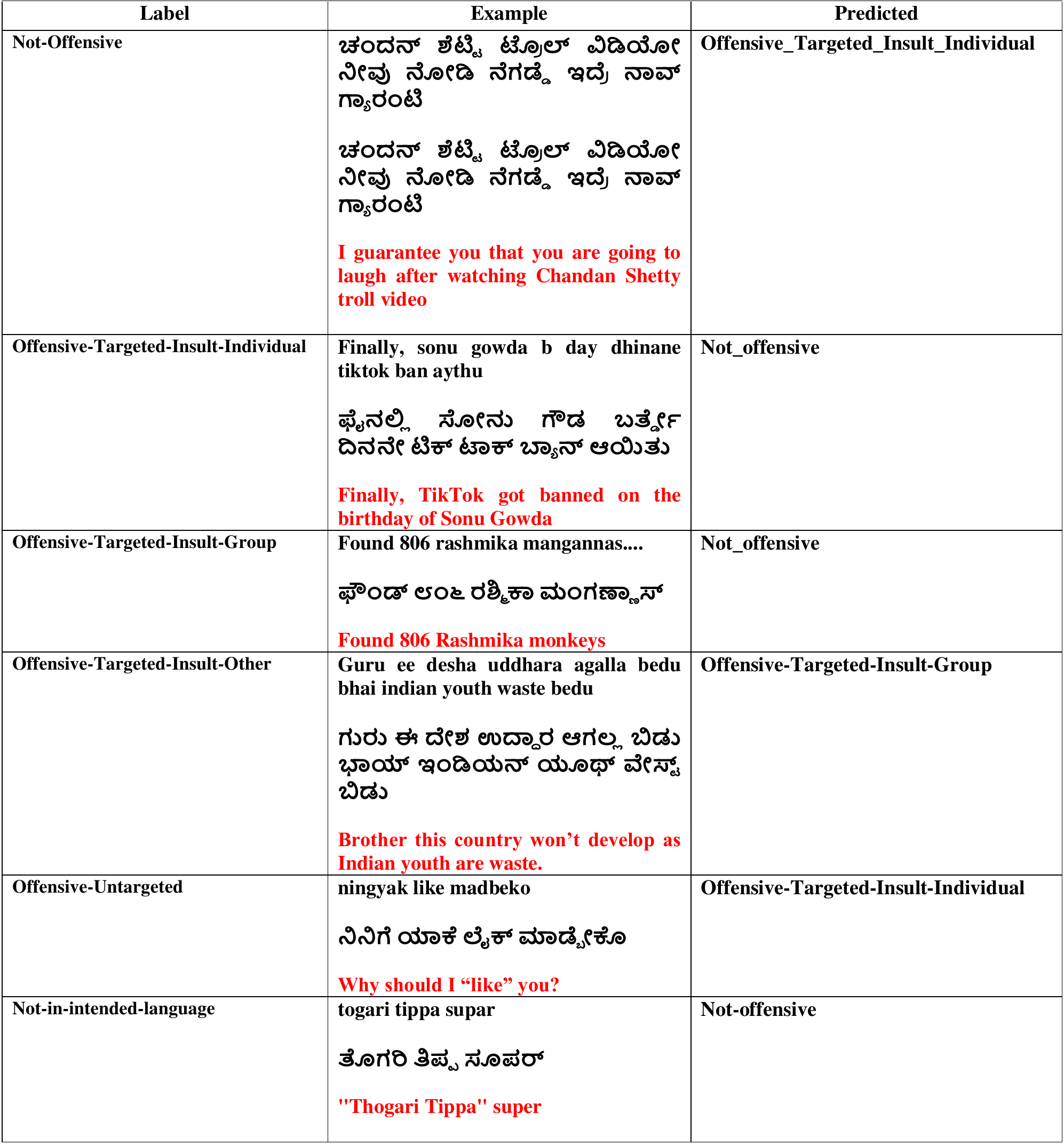}
    \caption{Wrong Predictions in the Kannada dataset}
    \label{fig:kan_error}
\end{figure}

\begin{figure}[htpb]
    \centering
    \includegraphics[width=0.8\linewidth, height=12cm]{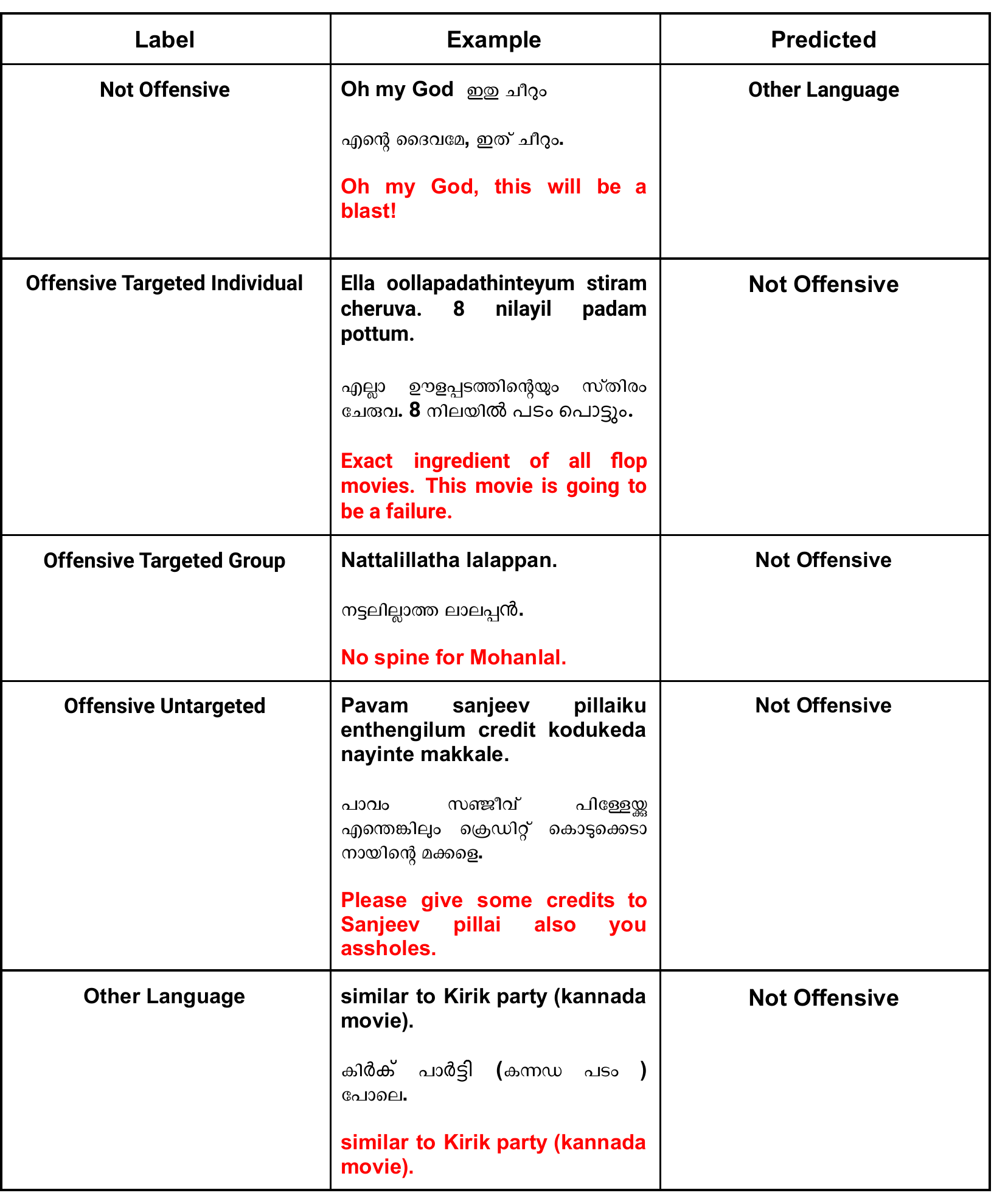}
    \caption{Wrong Predictions in the Malayalam dataset}
    \label{fig:mal_error}
\end{figure}
The results obtained for this task has succeeded in surpassing many of the state-of-the-art models on the dataset \cite{chakravarthi-etal-2021-findings-shared-task}. For Tamil, our model secured the first position with an F1-score of 0.79 by surpassing \cite{saha-etal-2021-hate} who scored an F1-score of 0.78 as observed in Table \ref{tab:comp_tam}. The model ranked third in Kannada after \cite{jayanthi-gupta-2021-sj} and \cite{saha-etal-2021-hate} with an F1-score of 0.73 while the first ranker scored 0.75 as shown in Table \ref{tab:comp_kan}. For Malayalam, the model came second with an F1-score of 0.9649 while the first rank is shared by 4 teams \cite{saha-etal-2021-hate}, \cite{balouchzahi-etal-2021-mucs}, \cite{kedia-nandy-2021-indicnlp} and \cite{tula-etal-2021-bitions} with 0.97 as seen in Table \ref{tab:comp_mal}. \cite{saha-etal-2021-hate} used a genetic algorithm technique for ensembling different transformer models. \cite{jayanthi-gupta-2021-sj} also adopted an ensembling method using mBERT and XLM-RoBERTa with a masked language modelling objective. The confusion matrices of the best performing models are displayed in Figure \ref{fig:kan}, Figure \ref{fig:mal}, and Figure \ref{fig:tam}.

\subsection{Error Analysis}
\subsubsection{Tamil}
In this section, we discuss the misclassification errors of the model for the classification task. Some of the examples with English translations are discussed for each target class in Fig. \ref{fig:tam_error}. Here, the most six misclassified comments are discussed. 1,002 comments out of 4,392 in the testing set are wrongly predicted, and 3,390 sentences are correctly predicted. Most of the wrong predictions are that the offensiveness is not projected explicitly in terms of words in such sentences. However, the emotions in those sentences depict the offensiveness. For example, consider the following sentence.

Text: \emph{\textcolor{red}{Epudi da oru ANJAN  oru AYAN oru NJK. Mari irukumo}}

Translation: \emph{How it would be? May be Anjan,Ayan or NJK?}\\
In this sentence, no offensive word is used. However, the sarcasm in the text is offensive, and it would be understood by the people who know the Tamil Cinema and who watched Anjan, Ayan and NJK movies already (All are Tamil movie titles).
Another problem in some classifications is that the sentences are made with positive words, yet it gives offensive meaning implicitly. The sentence is misclassified as not offensive, though it comes under the category of ‘Offensive-Untargeted’.  Consider another example,  

Text: \emph{\textcolor{blue}{Rajini political entry dailog 1996/2016  10 years one dailog} \textcolor{red}{naaku baag nachindi}}

Translation: \emph{Political entry dialog of Rajnikanth between 1996-2016. I liked it.}\\
This sentence also does not have any offensive words. Though it has positive words such as `I liked it’, One can only understand the sarcasm if one knows Tamil movies and the actor Rajnikanth’s dialogues in his movies. Hence, the given sentence is not predicted as offensive even though it comes under Offensive-Targeted-Insult-Other.

\subsubsection{Kannada}

The classifications in the Kannada language saw the poorest F1-scores of the three languages. The test dataset consisted of 778 sentences, out of which 552 sentences were classified well, while 226 sentences failed to be grouped into their respective class. The confusion matrix can be viewed in Figure \ref{fig:kan}. We can see that many sentences are mismatched for various reasons that the model cannot detect. Consider the following sentences,

Text: \emph{\textcolor{blue}{Finally sonu gowda b day} \textcolor{red}{dhinane} \textcolor{blue}{tiktok ban} \textcolor{red}{aythu}}

Translation: \emph{Finally, TikTok got banned on the birthday of Sonu Gowda}//
The sentence might sound like a fact. It does not contain any offensive words or anything. However, people who know Sonu Gowda (a TikTok star) would understand the sarcasm behind the comment on how TikTok was banned in India just on her birthday. Therefore, the model does not predict it as Offensive-Targeted-Insult-Individual, the class to which it should belong. 

Text: \emph{\textcolor{blue}{Found 806 rashmika} \textcolor{red}{mangannas....}}

Translation: \emph{Found 806 Rashmika monkeys}\\
This sentence is very confusing due to the use of puns by the commenter. Rashmika Mandanna is a well known Indian actress, and ``manganna" which sounds like ``Mandanna” means monkey in Kannada. This replaced alphabet can be considered a spelling error, and the sentence could be classified into other classes like "not-Kannada" or "Not-offensive". Hence, the model failed to categorize it into “Offensive-Targeted-Insult-Group”, which is the suitable class.

Text: \emph{\textcolor{red}{togari tippa supar}}

Translation: \emph{"Thogari Tippa" super}\\
"Thogari Tippa" is the title of a Kannada movie and is derived from Kannada. The model fails to detect that "Thogari Tippa" is a movie and classifies it into "Not-offensive". However, the sentence belongs to the group, "not-Kannada".

\subsection{Malayalam}
The performance of all the models on the code-mixed Malayalam dataset was very exceptional, as most of the models achieved weighted F1-Scores greater than 0.90. The best performing model misclassified relatively lesser examples in contrast to its performance on other datasets, misclassifying 78 samples among the 2,001 samples in the test set.

Text: \emph{\textcolor{red}{Ella oollapadathinteyum stiram cheruva.} \textcolor{blue}{8} 
\textcolor{red}{onilayil padam pottum}.}

Translation: \emph{Exact ingredient of all flop movies. This movie is going to be a failure.}\\
Though this comment does not have any offensive words, this sentence as a whole is meant to insult the director or any person behind that movie. Based on just  a trailer, the author sees the film with prejudice and gives negative reactions about the movie itself. Hence the sentence is wrongly classified as ‘Not Offensive’ but it belongs to the ‘Offensive Targeted Individual’ category. Let us see another example,

Text: \emph{\textcolor{red}{Valla panikkum pokkude nalla prayam indallo mammotty}}

Translation: \emph{You are too old Mammootty. Can’t you do some real work?}\\
This comment also does not contain any offensive words but this explicitly insults a prominent actor of Malayalam movies. Mammootty is considered one of the legends among Indian actors yet this comment deliberately humiliates him. So this comment with no doubt should fall into the ‘Offensive Targeted Individual’ category but wrongly predicted as ‘Not Offensive’ as there are no offensive words in the comment.

\section{Conclusion}
\label{conclusion}
The increasing amount of offensive language persisting on social media and relatively fewer approaches that address code-mixing in Dravidian languages have pushed us to improve our approaches in offensive language identification. This paper revisited offensive language identification by prioritising the dataset rather than the models to increase the overall weighted F1-scores on the test set. We intend to address the lack of data by generating pseudo-labels on the dataset, which was transliterated to the respective Dravidian languages, as Dravidian languages are primarily under-resourced. This approach aims at improving cross-lingual transfer by having a multilingual training dataset. In this paper, we have presented an approach to identify the offensive language in social media for multilingual countries that comprise code-mixed sentences. We observe that when we fine-tune ULMFiT on our newly constructed custom dataset, it yields the best performance in code-mixed Tamil while almost achieving very competitive results on several other benchmarked models on the respective languages. For future work, we intend to combine the three languages and develop a multilingual offensive language identification system for code-mixed Dravidian languages.

\printcredits
\section*{Acknowledgements}
The author Bharathi Raja Chakravarthi was supported in part by a research grant from Science Foundation Ireland (SFI) under Grant Number SFI/12/RC/2289$\_$P2 (Insight$\_$2), co-funded by the European Regional Development Fund and Irish Research Council grant IRCLA/2017/129 (CARDAMOM-Comparative Deep Models of Language for Minority and Historical Languages).
\bibliographystyle{cas-model2-names}

\bibliography{cas-refs}


\end{document}